\documentclass[final,5p,times,twocolumn,authoryear]{elsarticle}

\usepackage{amssymb}
\usepackage{amsmath}

\usepackage{amsfonts}
\usepackage{graphicx}
\usepackage{booktabs}
\usepackage{url}
\usepackage{silence}
\usepackage{subcaption}
\DeclareCaptionLabelFormat{boldalg}{\textbf{#1~#2}}
\usepackage{float} 
\usepackage{textcomp}
\usepackage{tcolorbox}
\usepackage{enumitem}
\usepackage{changepage}
\usepackage{fancyhdr}
\usepackage{hhline} 
\usepackage{multirow}
\usepackage{fontawesome5}
\usepackage{pifont}

\newcommand{\hcmark}{\textcolor{green!50!black}{\faCheck}}
\newcommand{\hxmark}{\textcolor{red}{\faTimes}}
\usepackage{colortbl}
\usepackage[table]{xcolor} 
\definecolor{lightgray}{gray}{0.9}
\definecolor{darkgray}{gray}{0.8}
\definecolor{darkgreen}{RGB}{0,100,0}
\definecolor{mygreen1}{RGB}{240, 245, 230}
\definecolor{mygblue1}{RGB}{217, 243, 253}
\definecolor{mygorange1}{RGB}{254, 239, 227}
\definecolor{mygreen2}{RGB}{92, 187, 86}
\definecolor{mygblue2}{RGB}{2, 183, 232}
\definecolor{mygorange2}{RGB}{247, 148, 29}
\usepackage[colorlinks=true, linkcolor=blue, citecolor=blue, urlcolor=blue]{hyperref} 

\usepackage[capitalize]{cleveref} 
\usepackage{makecell}
\usepackage{algorithm}
\usepackage{algorithmicx}
\usepackage{algpseudocode}
\usepackage{siunitx}
\journal{Engineering Applications of Artificial Intelligence}

\begin{document}

\begin{frontmatter}

\title{TailBooster: A Dual-Layer Generative Framework for Extreme Value Augmentation with Operational Validity Enforcement} 

\author{\texorpdfstring{Karim Aly\corref{cor1}}{Karim Aly}} 
\ead{k.y.s.b.aly@tudelft.nl}
\cortext[cor1]{Corresponding author}

\author{Alexei Sharpanskykh}
\ead{o.a.sharpanskykh@tudelft.nl}

\author{Jacco Hoekstra}
\ead{j.m.hoekstra@tudelft.nl}

\affiliation{organization={Control and Operations, Faculty of Aerospace Engineering, 
    Delft University of Technology (TU Delft)}, 
    addressline={Kluyverweg 1}, 
    city={Delft},
    postcode={2629 HS}, 
    state={South Holland},
    country={Netherlands}}

\begin{abstract}
Extreme events in air transportation, such as severe arrival delays and abnormal air times, cause cascading network disruptions with substantial operational, economic, and safety costs. Such events are rare in historical records, leaving insufficient training signal for machine learning models to learn their patterns. Synthetic data augmentation offers a principled solution, but conventional generative models under-represent distributional tails and give no guarantee against operationally infeasible instances, such as a short air time paired with a long flight distance. No existing approach addresses both limitations for mixed-type tabular records. We therefore propose \emph{TailBooster}, a dual-layer generative framework combining generative modelling with two anomaly detection layers. A statistical layer extracts extremes via the interquartile range, supplying tail-concentrated training signal to dedicated generative models, here a Tabular Variational Autoencoder. A deep learning layer then applies autoencoder-based cleaning, discarding synthetic records that violate the operational envelope learned from historical data. The framework was evaluated on U.S.\ flight records across five dimensions: diversity, statistical similarity, fidelity, operational validity, and utility, the latter two constituting the primary improvement targets. Data-driven cleaning markedly improved operational validity, while targeted augmentation enhanced utility for extreme-event prediction. Across six regression algorithms, training on the framework's records reduced Mean Absolute Error by 47--49\% on extreme air time and 29--57\% on extreme arrival delay prediction relative to conventional synthetic data, with comparable gains when real records were enriched with synthetic extremes. Being fully data-driven and model-agnostic, \emph{TailBooster} extends naturally to domains where extreme-event prediction is critical and domain-specific rules are unavailable.
\end{abstract}


\end{frontmatter}


\section{Introduction}
\label{sec:introduction}
Air transportation is periodically disrupted by extreme operational events, most notably severe arrival delays and abnormal air times, whose consequences extend well beyond the immediate incident, affecting airlines, airports, passengers, and air traffic management. A single extreme delay can trigger cascading disruptions across a network, incurring substantial costs in crew rescheduling, aircraft repositioning, and passenger compensation \citep{Cook2015}, while abnormal air times, whether caused by technical irregularities or adverse meteorological conditions, drive excess fuel consumption and compromise both operational efficiency and safety. Reliable prediction of such events is therefore essential for proactive decision-making and operational resilience.

In historical flight records, extreme events remain inherently rare relative to nominal operations, and this rarity creates a fundamental learning problem. The distributional tails of numerical features, such as air time and arrival delay, are sparsely populated in historical data, providing insufficient training signal for machine learning models to learn the patterns that characterise these tails. As air time and arrival delay are continuous operational metrics rather than discrete categories, predicting their magnitude is naturally formulated as a regression task. The result is regression models that generalise well across nominal conditions but fail precisely where accurate prediction matters most: at the extremes.

This challenge is compounded by limited public accessibility of historical flight data beyond the organisations owning it. Aviation stakeholders, such as airlines, airports, and air navigation service providers, hold their own historical records, but data-sharing among them remains constrained by commercial, regulatory, and privacy considerations \citep{Aly2025a}. Researchers and analysts operating outside these organisations therefore face a dual obstacle: not only are extreme events rare within any single dataset, but access to detailed and representative operational records is itself restricted. For instance, in Europe, the EUROCONTROL Demand Data Repository~(DDR), which contains historical traffic demand and Air Traffic Flow Management~(ATFM) delay data, is accessible only to licensed air navigation service providers and airline operators \citep{SynthAIr2025}. Similarly, European flight schedules are owned by individual airlines, and researchers seeking access can only purchase them from private agencies \citep{SynthAIr2025}. For this reason, the present study uses the domestic flight records from the U.S.\ Bureau of Transportation Statistics~(BTS)~\citep{TranStats, bts}, which represent the only publicly accessible source of per-flight operational records combining scheduled and actual times, flight status classifications (delayed, cancelled, and diverted) alongside their documented causes, and detailed diversion metadata, all at the route level.

Synthetic data generation offers a principled pathway to address both obstacles. By augmenting historical records with synthetic extremes, practitioners with access to real data can enrich the tail regions of their datasets and improve the predictability of extreme events; for those without such access, high-quality synthetic data can serve as a substitute, enabling model development and benchmarking without requiring direct access to historical records. However, beyond reproducing the overall distribution of individual features, standard generative models offer no guarantee of preserving the operational correlations between them, and may therefore generate records that are operationally infeasible, such as a short air time paired with a long flight distance, which makes these synthetic records unsuitable to train downstream task models or decision-making systems \citep{Aly2025a}. Additionally, they systematically under-represent distributional tails, irrespective of how well they approximate the overall distribution \citep{Huster2021, Bhatia2021, Gu2025}. Approaches that embed Extreme Value Theory~(EVT) into generative architectures to address the tail under-representation deficit \citep{Bhatia2021, Lafon2023} are designed for continuous, single-type feature spaces, such as geophysical fields or financial return series, and have not been extended to the mixed-type tabular records characteristic of aviation data. Conditional generation \citep{xu2019modeling} is applicable to mixed-type tabular data but does not resolve this deficit, since conditioning operates at sampling time rather than at training time: selecting a rare class as the conditioning label changes which records are retrieved from the model, but it does not alter the sparse tail-region signal the model received during training, which is the actual cause of the under-representation \citep{Khorram2024}. As a result, the conditionally sampled extremes cannot be guaranteed to be realistic or operationally valid. Outlier-focused tabular architectures address the mixed-type setting and generate outliers, but without targeting specific distributional tails of user-defined continuous features or enforcing operational validity \citep{Azimi2024}. Domain-constrained generative models do enforce operational plausibility, but through hand-crafted symbolic rules that require domain-specific re-engineering for each new application context and presuppose that the governing physical relationships are known, which is not always the case, such as for the correlation between air time and flight distance for specific airport pairs \citep{Karimanzira2024, Yi2024}. No existing approach therefore simultaneously targets tail under-representation in mixed-type tabular records and enforces operational validity in a fully data-driven fashion. 

The objective of this study is therefore twofold: to develop a generative framework that combines generative modelling with anomaly detection to produce synthetic flight records with improved extreme representativeness and operational validity, and to evaluate whether augmenting training data with such records translates into measurable gains in the predictability of extreme events, i.e., regression accuracy on extreme target values. 

To this end, this study proposes \emph{TailBooster}, a dual-layer generative framework for extreme-value augmentation with operational validity enforcement. The dual-layer design brackets a Tabular Variational Autoencoder (TVAE) generative stage---chosen for its comparative training stability and lower computational cost relative to alternative generative architectures---with two anomaly detection layers: a statistical layer that precedes generation and a deep learning layer that follows it. The framework takes as input the full historical dataset together with two user-defined feature lists: (i)~the target features whose distributional tails are to be augmented, used by the statistical layer to isolate extreme subsets, and (ii)~the operationally correlated features, i.e., features whose joint values are governed by operational constraints, used by the deep learning layer to characterise operational feasibility. The process starts with the statistical layer, which performs IQR-based extreme-subset extraction prior to generation, isolating one extreme subset per target feature to supply a tail-concentrated training signal to dedicated generative models. A TVAE is then trained on the full dataset and another TVAE on each extreme subset; synthetic records are then sampled from all models, and the resulting records are filtered to retain only those whose origin--destination airport pairs appear in the historical data. After generation, the deep learning layer applies pre-trained autoencoders to these candidate synthetic records, discarding samples that fall outside the empirical operational envelope learned from the historical data. This data-driven cleaning step enforces operational validity without requiring hand-crafted domain rules to be available. The pipeline yields three synthetic or augmented datasets---Na\"{i}ve Synthetic, Augmented Synthetic, and Augmented Real---which, together with the original Real dataset, are assessed across five complementary dimensions: diversity, statistical similarity, fidelity, operational validity, and regression utility on extreme subsets. This assessment adopts and extends the multi-dimensional evaluation framework from our earlier work \citep{Aly2026}. Although demonstrated here in an aviation context, the framework is transferable to any domain where the prediction of extreme values of target numerical features is operationally critical, with its fully data-driven operational cleaning making it particularly suitable for settings where physical governing equations or domain-specific symbolic rules for enforcing operational validity are unavailable. Finally, the generative component is model-agnostic: the TVAE can be replaced by any tabular generative model without altering any other part of the pipeline.

The contributions of this research are fourfold: (i)~a dual-layer generative framework for mixed-type tabular data that combines the representational capacity of deep generative models with statistical and deep learning anomaly detection, jointly addressing tail under-representation and enforcing operational validity in a fully data-driven manner; (ii)~a demonstration that the data-driven operational cleaning mechanism markedly improves the operational validity of synthetic flight records relative to conventional generation; (iii)~evidence that targeted extreme value augmentation consistently improves the predictability of extreme events across six models spanning three distinct regression families; and (iv)~a comparative evaluation demonstrating measurable quality gains both for practitioners with access to historical flight records, through augmentation of real data with operationally valid synthetic extremes, and for those without such access, through augmented and operationally cleaned synthetic data that outperforms conventionally generated synthetic data.

The remainder of this paper is organised as follows: Section~\ref{sec:related_work} reviews the literature. Section~\ref{sec:methodology} presents the \emph{TailBooster} architecture, evaluation framework, and hyperparameter tuning procedure. Section~\ref{sec:results} reports experimental results across all five evaluation dimensions. Section~\ref{sec:discussion} interprets key design choices and their practical and computational implications. Section~\ref{sec:conclusions_and_future_work} concludes with key findings and directions for future work.

\section{Related work}
\label{sec:related_work}

The literature relevant to this work spans five interconnected areas. We first situate the proposed framework within the broader landscape of generative models for tabular data synthesis, establishing the architectural baseline and its well-documented tail blindspot (Section~\ref{sec:related_work_theme1}). We then introduce Extreme Value Theory~(EVT) as the statistical foundation for principled tail characterisation (Section~\ref{sec:related_work_theme2}), before reviewing the family of EVT-augmented generative models that embed this theory into deep learning architectures (Section~\ref{sec:related_work_theme3}). The review subsequently narrows to outlier-focused architectures for tabular data, with particular attention to \emph{zGAN} as the closest existing implementation to \emph{TailBooster} (Section~\ref{sec:related_work_theme4}), and closes with domain-constrained generative models that enforce operational plausibility through explicit physical constraints or implicit conditioning (Section~\ref{sec:related_work_theme5}). 

\subsection{Generative Models for Synthetic Tabular Data}
\label{sec:related_work_theme1}

Over the past decade, generative modelling has emerged as a leading paradigm for synthesising tabular data in domains where data scarcity, privacy constraints, or class imbalance limit the direct use of historical records. The literature is dominated by three architectural families: generative adversarial networks, variational autoencoders, and diffusion models. Generative adversarial networks (GANs) \citep{goodfellow2014generative}, in which a generator and discriminator are trained in a minimax adversarial loop, were among the first to demonstrate high-fidelity synthesis of complex distributions, though their adversarial training dynamics remain prone to instability and mode collapse \citep{trilemma}. Their tabular adaptations, most notably the Conditional Tabular GAN \citep{xu2019modeling} and the Copula GAN \citep{sdv}, introduced mode-specific conditional sampling and copula-based dependency modelling to accommodate the mixed data types and complex dependency structures characteristic of tabular records. Variational autoencoders (VAEs) \citep{Kingma2013} and their tabular adaptation, the Tabular Variational Autoencoder (TVAE) \citep{xu2019modeling}, introduced a complementary probabilistic framework in which an encoder maps observations into a regularised latent distribution and a decoder reconstructs them by maximising the evidence lower bound (ELBO) on the data likelihood, enabling training that is less susceptible to the mode collapse affecting GANs and computationally lighter than diffusion models \citep{trilemma}. More recently, diffusion models \citep{DIFF_intro}, which learn to reverse a progressive noise-corruption process through iterative denoising, have shown strong training stability and sample diversity in image and continuous-data domains, albeit at a substantially higher computational cost due to their iterative sampling process \citep{trilemma}; their application to tabular data is comparatively recent, though emerging work suggests they may offer advantages in capturing complex multivariate dependencies \citep{Kotelnikov2023, Gu2025}.

These architectures have been applied to aviation data synthesis in our prior work. \citet{Aly2025a} demonstrated their ability to generate synthetic flight records of sufficient statistical fidelity to support arrival delay prediction, using a dataset consisting exclusively of nominal operations without extreme events, while \citet{Aly2026} extended this work to augment rare flight diversion records through a multi-objective hyperparameter optimisation scheme, showing that targeted synthetic augmentation could measurably improve the predictive performance of downstream classifiers. Across both studies, the generated data faithfully captured the central tendencies and common operational patterns observed in the historical distributions.

However, all three architectural families share a structural limitation that directly motivates the present work. Their training objectives, whether adversarial, ELBO maximisation, or denoising score matching, inherently prioritise the high-density regions of the data distribution, where most of the training signal resides. Extreme observations residing in the distributional tails are consequently under-represented in the generated data, irrespective of how well the overall distribution is approximated \citep{Huster2021, Bhatia2021, Lafon2023, Gu2025}. Conditional generation, where the model is conditioned on a class label or feature value during sampling, appears to offer a solution. In practice, however, conditioning cannot compensate for a lack of representative training examples. When the conditioned class is rare, the model receives insufficient signal to learn that region of the distribution, often producing low-fidelity or degenerate samples rather than resolving the tail coverage deficit \citep{Khorram2024}. This tail blindspot is therefore a direct consequence of what these models are trained to optimise: accuracy over the bulk of the distribution, with no incentive to represent its tails. Addressing this limitation requires a statistical framework that explicitly characterises tail behaviour and provides a principled basis for modelling extremes. Extreme Value Theory provides this foundation.

\subsection{Extreme Value Theory for Tail Modelling}
\label{sec:related_work_theme2}

Extreme Value Theory (EVT) provides the statistical foundation for characterising the behaviour of distribution tails and has long served as the principled framework for modelling rare, high-impact events in engineering, finance, and environmental science \citep{deHaan2006, Coles2001, Embrechts1997}. Its central theoretical result states that, under mild regularity conditions, the distribution of extreme observations converges to one of a small family of limiting distributions, a result that holds across a broad class of underlying data distributions, making EVT well suited to applications where the parent distribution is unknown.

Two complementary methods operationalise this result: the Block Maxima method, which models the maximum value within non-overlapping blocks of observations \citep{Ferreira2015}, and the Peaks-Over-Threshold (POT) method, which models all observations exceeding a user-defined threshold via the Generalised Pareto Distribution (GPD) \citep{Leadbetter1991, Pickands1975, Balkema1974}. For continuous data in which extreme events occur sporadically, POT is generally preferred, as it makes more efficient use of the available observations than block-based aggregation \citep{Coles2001}. Fitting a GPD to the empirical tail yields a model capable of extrapolating beyond the largest observed value, a capability that standard generative models and simple resampling strategies generally lack. This treatment of the tail as a distinct statistical regime motivated our IQR-based extreme subset extraction, a non-parametric, operationally driven analogue to POT thresholding: rather than fitting a parametric GPD, we isolate the empirical tail directly and use it as a targeted training signal for the generative component of \emph{TailBooster}.

EVT's extrapolative power, however, is confined to a single feature at a time: EVT is a tail estimator, not a generator, and on its own cannot produce synthetic multivariate records that preserve inter-feature dependencies, heterogeneous data types, and operational constraints. Realising its benefits for data synthesis has therefore motivated embedding EVT principles into generative models, giving rise to a growing body of EVT-augmented methods, reviewed next.

\subsection{EVT-Augmented Generative Models for Extreme Synthesis}
\label{sec:related_work_theme3}

Building upon this statistical foundation, EVT principles have been integrated directly into generative models to improve the synthesis of extreme observations: through modifications to the training objective, at the sampling or conditioning stage, or within the latent representation.

The first strategy embeds EVT constraints directly into the training objective, as in \emph{Pareto GAN} \citep{Huster2021}, which incorporates a tail-index-aware loss to stabilise training on heavy-tailed marginals, and \emph{EV-GAN} \citep{Allouche2022}, which reparametrises the generator to better approximate unbounded quantiles. The second strategy conditions the generator on EVT-fitted tail distributions: \emph{ExGAN} \citep{Bhatia2021} conditions generation on a GPD-derived extremeness measure to sample at a user-specified extremeness probability, and its spatial extension \emph{evtGAN} \citep{Boulaguiem2022} captures tail dependence across geographically distributed climate variables. The third strategy modifies the latent representation to encode heavy-tailed structure: \emph{HTGAN} \citep{Girard2025} replaces the standard Gaussian latent prior with a heavy-tailed distribution, while \emph{ExtVAE} \citep{Lafon2023} applies multivariate regular variation theory to model event magnitude and dependence structure separately via a radius--angle decomposition.

Taken together, these contributions represent a substantial advance over standard generative baselines. Yet a common limitation remains: existing EVT-augmented generative methods have primarily been developed for continuous univariate or multivariate data, such as geophysical fields, climate records, and financial returns, and generally assume continuous, homogeneous feature spaces amenable to probabilistic reparametrisation. This assumption breaks down for mixed-type tabular records, which often combine categorical, discrete numerical, continuous numerical, and datetime features, as in aviation, where such heterogeneous variables must be modelled jointly while preserving domain-specific validity.

\subsection{Outlier-Focused Generative Architectures for Tabular Extreme Augmentation}
\label{sec:related_work_theme4}

A parallel strand of the literature has addressed outlier and rare-event synthesis directly in the tabular domain, motivated primarily by data imbalance in financial applications. \citet{Jiang2024} represents this strand, proposing GAN-based augmentation schemes that generate synthetic minority-class instances, corresponding to rare systemic risk events or extreme market conditions, to rebalance training datasets and improve downstream classification performance. This work demonstrated that adversarial augmentation of rare tabular records can improve minority-class detection metrics, yet it does not enforce structural constraints ensuring that the generated records are operationally coherent beyond their distributional similarity to the training data.

The most explicitly outlier-focused treatment is \emph{zGAN} \citep{Azimi2024}, developed and validated on financial services data, which is also the closest existing implementation to the framework proposed in this study. \citet{Azimi2024} integrated a Conditional Variational Autoencoder (CVAE) and a GAN within a unified pipeline. The CVAE learns structured latent representations of real tabular data and generates synthetic covariance matrices that encode inter-feature dependencies. These matrices are then passed to a dedicated Outlier Conditional Covariance Generator (\emph{covGEN}) to sample macro-outliers. Outlier magnitudes are drawn from heavy-tailed distributions, including Laplace, Weibull, Gumbel, and Lévy distributions, parameterised by the estimated covariance. A hash-based similarity filter subsequently removes synthetic records that closely resemble real training instances, thereby preserving privacy. The final output combines baseline synthetic data with outlier-augmented records, while a Target Model module based on gradient-boosted classification imputes predicted target labels into the synthetic records.

Despite these contributions, zGAN exhibits three limitations that are consequential for the present study. First, its evaluation framework is oriented towards binary classification utility, assessed using the Area Under the Curve metric, rather than towards regression accuracy on extreme target values, which is the primary objective here. Second, outlier magnitude is governed by covariance geometry, with no explicit mechanism for verifying that the generated records satisfy operational validity constraints beyond the statistical structure of the training data. Third, the architecture is GAN-dependent and designed for financial tabular data, leaving open the question of how domain-specific operational validity can be enforced in aviation, where feasibility constraints are often relational rather than purely statistical. This limitation points directly to a broader design challenge that domain-constrained generative models have sought to address.

\subsection{Domain-Constrained Generative Modelling and Operational Validity}
\label{sec:related_work_theme5}

The outlier-focused architectures reviewed in the preceding section demonstrate that extreme tabular synthesis is feasible, yet they share a common assumption: that statistical similarity to the training distribution is sufficient to ensure the usefulness of the generated records. In engineering and environmental domains, this assumption often fails. A synthetic flood record with statistically plausible discharge values is not useful if it violates mass conservation; a synthetic power-grid scenario is not useful if generation and demand do not balance. Extreme events in such settings must therefore respect the physical, operational, or systemic constraints that govern the real system, not merely resemble its historical statistics. This requirement has motivated a distinct strand of work in which domain knowledge is embedded directly into the generative process.

In the hydrological domain, \emph{MC-TSGAN} \citep{Karimanzira2024} embeds mass conservation, energy balance, and hydraulic continuity as regularisation terms in a time-series GAN's loss function, keeping synthetic extreme flood-event sequences physically consistent and improving multi-step runoff forecasting. In the energy domain, the same principle of constraint embedding is pursued through conditional architectures: an improved \emph{InfoGAN} variant \citep{Yi2024} preserves the joint correlation between wind, solar, and load variables under extreme demand conditions via mutual-information-based latent structure, while \emph{C-DCGAN} \citep{Li2023} combines Wasserstein loss with a ranked selection mechanism to retain only scenarios satisfying extreme risk criteria across generation, storage, and demand. A complementary strategy, pursued in climate and geophysical applications, embeds domain consistency implicitly through conditioning: \emph{DiffESM} \citep{Bassetti2024} conditions a diffusion model on Earth System Model means to emulate extreme temperature and precipitation sequences with realistic spatio-temporal coherence, while \citet{Peard2023} combines GAN-based generation with EVT-based tail extrapolation to produce spatially coherent compound hazard scenarios. Across these examples, operational constraints are formulated as symbolic regularisers, conditional inputs, or post-hoc filters incorporated into the generative pipeline at training or sampling time.

Collectively, these approaches show that integrating domain knowledge is a powerful means of generating operationally credible extreme records in constrained systems, particularly where the governing physical laws are well established. Their applicability, however, is bounded by this same requirement: governing laws such as mass-conservation terms, grid-stability conditions, and physical parameterisations are domain-specific, limiting their transferability to other operational settings. Additionally, they presuppose that such laws are known and formalisable in the first place, which is not always the case. This motivates a data-driven alternative capable of learning operational plausibility empirically from historical records rather than encoding it symbolically, particularly for settings where governing laws are unknown. In \emph{TailBooster}, this role is fulfilled by the autoencoder-based anomaly cleaning layer.

The foregoing review identifies four compounding gaps in the existing literature. First, standard generative architectures concentrate probability mass near the centre of the distribution, thereby systematically under-representing distributional tails \citep{Huster2021, Bhatia2021, Gu2025}. Second, although EVT provides the statistical foundation needed to characterise these tails, EVT-augmented generative models that build on it generally assume continuous, homogeneous feature spaces incompatible with mixed-type tabular aviation records \citep{Bhatia2021, Girard2025, Lafon2023}. Third, outlier-focused tabular architectures such as \emph{zGAN} \citep{Azimi2024} do not explicitly enforce operational validity and primarily target classification rather than regression utility. Finally, domain-constrained approaches embed operational knowledge through hand-crafted physical priors that are difficult to transfer across domains \citep{Karimanzira2024, Yi2024}.
The proposed \emph{TailBooster} responds to these gaps through a dual-layer, generative-model-agnostic augmentation framework that brackets tail-focused generation between an IQR-based statistical anomaly detection layer and an autoencoder-based anomaly cleaning layer, while incorporating a relational validity filter adapted to flight records. Evaluation adopts and extends the multi-dimensional framework developed in \citet{Aly2026} by combining standard distributional metrics with regression utility on extreme subsets, thereby assessing both tail coverage and operational validity. The architecture, training procedure, and evaluation protocol are described in full in the following section.

\section{Methodology}
\label{sec:methodology}
This section describes the data sources, preprocessing steps, and feature engineering decisions underpinning the study. It then presents the architecture of the proposed \emph{TailBooster} model. Finally, it details the evaluation criteria used to assess the quality of the generated data and the multi-objective hyperparameter tuning procedure applied to the generative stage.

\subsection{Data and Preprocessing}
\label{sec:data}
Flight records for this study were drawn from the TranStats Database for Airline On-Time Performance \citep{TranStats, bts}, a publicly accessible repository maintained by the Bureau of Transportation Statistics (BTS). The database compiles U.S.\ domestic flight information encompassing operational delay types, cancellations, and diversions, making it well suited to a broad range of air transport research applications.

Building upon our previous work \citep{Aly2025a}, which established the preprocessing and feature engineering pipeline, this study applies the same procedure to all arrivals and departures in New York State during January 2023, without restricting the scope to flights between two specific airports. Cancelled and diverted flights were excluded, as the present study targets the distributional tails of continuous on-time performance metrics, such as ``Air Time (min)'' and ``Arrival $\Delta$T (min)'', rather than the diversion minority class addressed in \citet{Aly2026}. The resulting dataset comprised 30~features and approximately 61,000 flight records, spanning 113~airports across 508~routes.

\begin{figure}[H]
    \centering
    \includegraphics[width=0.48\textwidth]{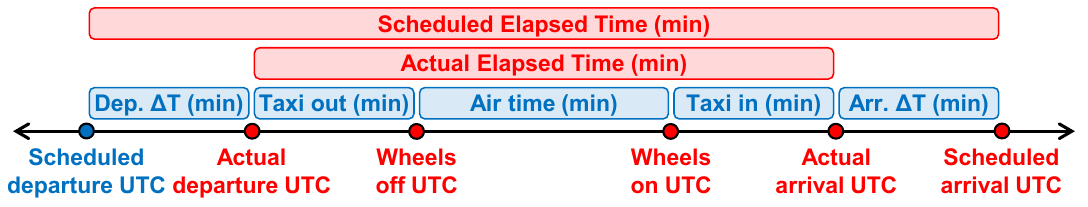}
    \caption{Temporal features used as direct inputs to the generative model (blue) and computed after generation (red).}
    \label{fig:temporal_features}
\end{figure}

To preserve temporal consistency between synthetic records, the features supplied to the generative models were required to be mutually independent, with no overlapping information between them. Figure~\ref{fig:temporal_features} illustrates the full set of temporal features in the historical records, distinguishing those used as direct inputs to the generative model from those derived after generation. Notably, ``Air Time (min)'' and ``Arrival $\Delta$T (min)'' were deliberately retained as generative model inputs rather than derived post-generation: both are continuous operational metrics whose distributional tails are the primary focus of this study, and both are designated throughout as \emph{target features}, the features whose extreme values \emph{TailBooster} is designed to augment in order to improve their predictability in downstream regression tasks. Preventing information overlap in the temporal features fed to the generative model ensures that derived features computed post-generation remain internally consistent by construction: for instance, the departure delay is guaranteed to equal the difference between the scheduled and actual departure times, air time to equal the interval between wheels-off and wheels-on, and arrival delay to equal the difference between the scheduled and actual arrival times. Additionally, priority was given to numerical duration features, expressed in minutes, over raw datetime features: datetime variables are difficult for machine learning models to handle directly and require either Unix epoch encoding, which discards cyclic and seasonal structure, or decomposition into multiple sub-fields, each of which increases the dimensionality of the generative model's input space and risks degrading synthesis quality. 

To minimise feature redundancy, 10~features were selected as direct inputs to the generative models: 6~temporal variables illustrated in Figure~\ref{fig:temporal_features} and 4~categorical features; the remaining attributes, designated as \emph{relational features}, were derived post-generation to preserve inter-feature dependencies and enhance the realism of the synthetic records. Table~\ref{tab:features} lists all features, indicating whether each was used for generation, and specifying the subsets employed for predicting ``Air Time (min)'' and ``Arrival $\Delta$T (min)'', as described in Section~\ref{sec:utility_methodology}.

\begin{table}[H]
  \centering
  \footnotesize
  \resizebox{\columnwidth}{!}{%
    \begin{tabular}{@{}lccc@{}}
      \toprule
      \textbf{Feature} &
      \makecell[c]{\textbf{Generation} \\ \textbf{input}} &
      \makecell[c]{\textbf{Prediction of} \\ \textbf{``Air Time (min)''}} &
      \makecell[c]{\textbf{Prediction of} \\ \textbf{``Arrival $\Delta$T (min)''}} \\
      \midrule
      Unique Carrier Code              & \hcmark       & \hcmark                           & \hcmark \\
      Tail Number                      & \hcmark       & \hcmark                           & \hcmark \\
      Origin Airport ID                & \hcmark       & \hxmark                           & \hxmark \\
      ICAO Origin Airport              & \faCalculator & \hcmark                           & \hcmark \\
      Origin City                      & \faCalculator & \hxmark                           & \hxmark \\
      Origin State Code                & \faCalculator & \hxmark                           & \hxmark \\
      Origin State Name                & \faCalculator & \hxmark                           & \hxmark \\
      Destination Airport ID           & \hcmark       & \hxmark                           & \hxmark \\
      ICAO Destination Airport         & \faCalculator & \hcmark                           & \hcmark \\
      Destination City                 & \faCalculator & \hxmark                           & \hxmark \\
      Destination State Code           & \faCalculator & \hxmark                           & \hxmark \\
      Destination State Name           & \faCalculator & \hxmark                           & \hxmark \\
      Quarter                          & \faCalculator & \hcmark                           & \hcmark \\
      Day of Week                      & \faCalculator & \hcmark                           & \hcmark \\
      Scheduled Departure Time UTC     & \hcmark       & \hcmark                           & \hcmark \\
      Actual Departure Time UTC        & \faCalculator & \hcmark                           & \hcmark \\
      Departure $\Delta$T (min)        & \hcmark       & \hcmark                           & \hcmark \\
      Departure Delay Label            & \faCalculator & \hxmark                           & \hxmark \\
      Taxi Out Time (min)              & \hcmark       & \hcmark                           & \hcmark \\
      Wheels Off Time UTC              & \faCalculator & \hcmark                           & \hcmark \\
      Wheels On Time UTC               & \faCalculator & \hxmark                           & \hxmark \\
      Taxi In Time (min)               & \hcmark       & \hxmark                           & \hxmark \\
      Scheduled Arrival Time UTC       & \faCalculator & \hcmark                           & \hcmark \\
      Actual Arrival Time UTC          & \faCalculator & \hxmark                           & \hxmark \\
      Arrival $\Delta$T (min)          & \hcmark       & \hxmark                           & \textcolor{blue}{\textbf{Target}} \\
      Arrival Delay Label              & \faCalculator & \hxmark                           & \hxmark \\
      Scheduled Elapsed Time (min)     & \faCalculator & \hcmark                           & \hcmark \\
      Actual Elapsed Time (min)        & \faCalculator & \hxmark                           & \hxmark \\
      Air Time (min)                   & \hcmark       & \textcolor{blue}{\textbf{Target}} & \hxmark \\
      Distance (miles)                 & \faCalculator & \hcmark                           & \hcmark \\
      \bottomrule
    \end{tabular}%
  }
  \caption{Features used in this analysis, categorised as: included
  (\hcmark), excluded (\hxmark), or calculated post-generation
  (\faCalculator).}
  \label{tab:features}
\end{table}

\subsection{Model Architecture}
\label{sec:model_architecture}
\renewcommand{\thesubsubsection}{\Alph{subsubsection}}

\begin{figure*}[t]
    \centering
    \includegraphics[width=\textwidth]{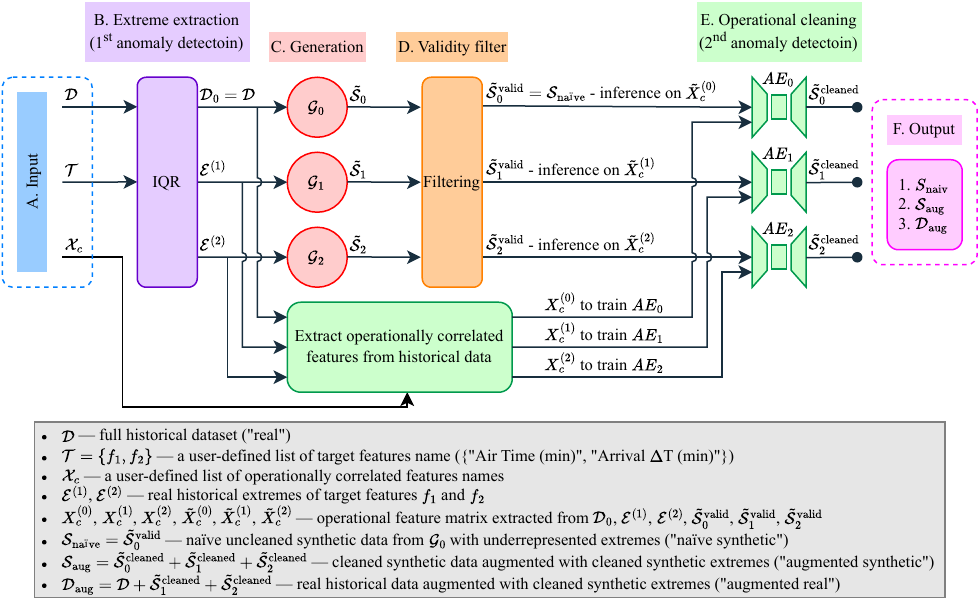} 
    \caption{Overview of the TailBooster architecture with two target features.}
    \label{fig:architecture}
\end{figure*}

Having described the dataset and feature engineering decisions, this subsection presents the architecture of the proposed \emph{TailBooster} framework. Figure~\ref{fig:architecture} provides an overview of the full pipeline. The framework processes historical flight records through a sequential pipeline comprising two anomaly detection layers, one statistical and one learned from historical data, with a generative stage and a relational validity filter interposed between them. The first anomaly detection layer addresses the well-known tendency of generative models to under-represent distributional tails by supplying tail-concentrated training signal, while the second corrects the complementary failure mode whereby generative models produce records that are statistically plausible but operationally invalid. Together, these components produce augmented synthetic data with improved extreme representativeness and operational validity. Algorithm~\ref{alg:tailbooster} provides a concise formal summary of the full pipeline, whose components are detailed in the remainder of this subsection.

\subsubsection{Input}
\emph{TailBooster} takes three inputs: the full historical dataset, hereafter referred to as the Real dataset and denoted $\mathcal{D} = \{x^{(i)}\}_{i=1}^{N}$, where $N$ is the number of historical records and $x^{(i)}$ denotes a single flight record; a user-defined list of target features $\mathcal{T} = \{f_1, \ldots, f_{N_{\mathrm{tf}}}\}$, that is, continuous operational metrics whose distributional tails are of interest; and a user-defined list of operationally correlated feature names $\mathcal{X}_c$, used to train the autoencoder-based anomaly cleaning layer. In the present study, the historical dataset is the U.S.\ domestic flight record described in Section~\ref{sec:data}, the target features are $\mathcal{T} = \{$``Air Time (min)'', ``Arrival $\Delta$T (min)''$\}$, giving $N_{\mathrm{tf}} = 2$, and $\mathcal{X}_c = \{$``ICAO Origin Airport'', ``ICAO Destination Airport'', ``Air Time (min)'', ``Distance (miles)''$\}$. The pipeline produces three datasets from these inputs, described in Section~\ref{sec:output} below, which are subsequently used together with the original Real dataset $\mathcal{D}$ to assess the usefulness of this framework.

\subsubsection{IQR-based extreme subset extraction}
This stage constitutes the first of the two anomaly detection layers in the pipeline. It partitions $\mathcal{D}$ according to the distributional tails of each target feature $f_j \in \mathcal{T}$. For each $f_j$, let $Q_1^{(j)}$ and $Q_3^{(j)}$ denote its first and third quartiles over $\mathcal{D}$. The interquartile range is defined as:

\begin{equation}
    \label{eq:iqr}
    \mathrm{IQR}^{(j)} = Q_3^{(j)} - Q_1^{(j)}
\end{equation}

The extreme subset for target feature $f_j$ is then the collection of records whose values fall outside the Tukey fences \citep{Tukey1977}:

\begin{equation}
    \label{eq:extreme_subset}
    \mathcal{E}^{(j)} = \left\{ x \in \mathcal{D} \;\middle|\;
    \begin{array}{l}
        x_{f_j} < Q_1^{(j)} - 1.5 \cdot \mathrm{IQR}^{(j)} \\[4pt]
        \text{or} \\[4pt]
        x_{f_j} > Q_3^{(j)} + 1.5 \cdot \mathrm{IQR}^{(j)}
    \end{array}
    \right\}
\end{equation}
where $x_{f_j}$ denotes the value of feature $f_j$ in record $x$. This operation is performed independently for each target feature, yielding the collection $\mathbb{D} = \{\mathcal{D}_0,\, \mathcal{E}^{(1)},\, \ldots,\, \mathcal{E}^{(N_{\mathrm{tf}})}\}$ of $N_{\mathrm{tf}} + 1$ datasets, where $\mathcal{D}_0 = \mathcal{D}$: one extreme subset per target feature plus the full historical record, giving three datasets in total. In the present study, records with ``Air Time (min)'' values outside $[-70.00,\, 282.00]$ minutes and ``Arrival $\Delta$T (min)'' values outside $[-65.50,\, 58.50]$ minutes are identified as extremes and assigned to $\mathcal{E}^{(1)}$ and $\mathcal{E}^{(2)}$, respectively. 

The IQR-based criterion was selected because it isolates distributional tails without imposing parametric assumptions on the underlying distribution, making it well suited to the mixed-type, heavy-tailed feature space characteristic of operational flight records. The collection $\mathbb{D}$ is subsequently used to train the generative models described in Section~\ref{sec:generative_training}.

\subsubsection{Generative model training}
\label{sec:generative_training}
One generative model $\mathcal{G}_k$ is trained on each dataset $\mathcal{D}_k \in \mathbb{D}$: one on the full historical record $\mathcal{D}_0$ to preserve nominal patterns, and one on each extreme subset $\mathcal{E}^{(k)}$ for $k = 1,\ldots,N_{\mathrm{tf}}$ to amplify the tail-region training signal for the corresponding target feature. In the present study, the Tabular Variational Autoencoder (TVAE) \citep{xu2019modeling} is used as the generative model. The TVAE extends the conventional variational autoencoder \citep{Kingma2013} to mixed-type tabular data by pairing a probabilistic encoder $q_\phi(z \mid x)$, which maps each input record $x$ to a distribution over a continuous latent space $\mathcal{Z}$, and a decoder $p_\theta(x \mid z)$ that reconstructs the record from a sampled latent vector $z$. Training maximises the Evidence Lower Bound (ELBO) on the marginal log-likelihood of the data:

\begin{align}
    \label{eq:elbo}
    &\mathcal{L}_{\mathrm{ELBO}}(\theta, \phi;\, x) = \nonumber \\
    &\mathbb{E}_{z \sim q_\phi(z \mid x)}\!\left[\log p_\theta(x \mid z)\right]
    - D_{\mathrm{KL}}\!\left(q_\phi(z \mid x) \,\|\, p(z)\right)
\end{align}

where the first term encourages accurate reconstruction of the input from its latent representation, and the Kullback--Leibler (KL) divergence term \citep{Kingma2013} regularises the approximate posterior $q_\phi(z \mid x)$ towards the prior $p(z)$, ensuring a well-structured and continuous latent space from which new records can be sampled \citep{Shen2024TowardsAF}. 

The TVAE was selected for this study given its training stability, reliable convergence, and reduced susceptibility to mode collapse relative to GAN-based architectures, as well as its lower computational cost compared to diffusion-based models. The framework is nonetheless generative-model-agnostic: the TVAE can be substituted with a conditional GAN or any other tabular generative model without modifying any other component of the pipeline.

Once all generative models are trained, records are sampled independently from each of them to produce a synthetic dataset. The sampling ratio $r = 1.2$ was set to compensate for the records expected to be removed during the subsequent post-generation filtering stages, ensuring that the final synthetic dataset retains a sufficient number of records relative to the original training data. Since $\mathcal{G}_0$ is trained on the full historical record, its synthetic output $\tilde{\mathcal{S}}_0$ is subject to the well-known tendency of deep generative models to concentrate probability mass in high-density regions of the data distribution, resulting in poor representation of the less frequent extreme observations of the target features $\mathcal{T}$ relative to the dominant nominal cases \citep{Huster2021, Bhatia2021}. In contrast, each $\mathcal{G}_k$ for $k = 1,\ldots,N_{\mathrm{tf}}$ produces synthetic extremes $\tilde{\mathcal{S}}_k$ corresponding to $\mathcal{E}^{(k)}$. Each synthetic dataset is then passed separately through the relational validity filter (Section~\ref{sec:validity_filter}) and the autoencoder-based operational cleaning layer (Section~\ref{sec:operational_cleaning}). 

\subsubsection{Relational validity filter}
\label{sec:validity_filter}
Before operational cleaning is applied, each synthetic record from the previous stage is subjected to a rejection sampling step that removes records whose origin--destination airport pair is absent from the historical data. Formally, let $\mathcal{P}_{\mathcal{D}} = \{(o^{(i)},\, d^{(i)}) : x^{(i)} \in \mathcal{D}\}$ be the set of all origin--destination pairs observed in the historical records. A synthetic record $\tilde{x}$ with origin airport $\tilde{o}$ and destination airport $\tilde{d}$ is accepted if and only if $(\tilde{o},\, \tilde{d}) \in \mathcal{P}_{\mathcal{D}}$, and discarded otherwise. This ensures that all remaining records correspond to operationally plausible routes observed in the historical data. Applying this filter to each $\tilde{\mathcal{S}}_k$ independently yields the corresponding validity-filtered datasets $\tilde{\mathcal{S}}^{\mathrm{valid}}_k$ for $k = 0,\ldots,N_{\mathrm{tf}}$, which are subsequently passed to the operational cleaning layer in Section~\ref{sec:operational_cleaning}. In particular, $\tilde{\mathcal{S}}^{\mathrm{valid}}_0$ is hereafter denoted $\mathcal{S}_{\text{na\"{i}ve}}$ and referred to as the Na\"{i}ve Synthetic dataset, representing the baseline output of a conventional generative approach trained on the full historical record, without operational cleaning or targeted extreme value augmentation.

\subsubsection{Autoencoder-based operational cleaning}
\label{sec:operational_cleaning}
This stage constitutes the second anomaly detection layer. It operates in two phases: a training phase, in which one autoencoder is trained once per dataset in $\mathbb{D}$ to learn the operational correlations between features in real historical records, and an operational cleaning phase, in which the pre-trained autoencoders are applied to remove synthetic records that violate empirically learned operational norms.

For each dataset $\mathcal{D}_k \in \mathbb{D}$, the operationally correlated feature matrix $X_c^{(k)}$ is extracted from $\mathcal{D}_k$ using the user-defined feature names $\mathcal{X}_c$. A standard autoencoder \citep{Hinton2006}, comprising an encoder $f_\phi^{(k)}$ and a decoder $g_\theta^{(k)}$, is then trained on $X_c^{(k)}$ by minimising the mean squared reconstruction loss over its $|X_c^{(k)}|$ records:

\begin{equation}
    \label{eq:ae_loss}
    \mathcal{L}_{\mathrm{AE}}(\theta^{(k)}, \phi^{(k)}) =
        \frac{1}{|X_c^{(k)}|} \sum_{x_c^{(i)} \in X_c^{(k)}}
        \left\| x_c^{(i)} - g_{\theta}^{(k)}\!\left(f_{\phi}^{(k)}
        \!\left(x_c^{(i)}\right)\right)
        \right\|_2^2
\end{equation}
where $x_c^{(i)}$ denotes the operational feature vector of record $i$ in $X_c^{(k)}$. By learning to reconstruct these features jointly, each autoencoder internalises the inter-feature constraints that characterise valid flight operations within its training distribution, for instance the proportionality between route distance and air time, without requiring domain-specific symbolic encoding. For each $k$, once $(f_\phi^{(k)}, g_\theta^{(k)})$ is trained, the per-record reconstruction error is computed as:

\begin{equation}
    \label{eq:ae_error}
    e^{(k)}(x_c) = \left\| x_c - g_\theta^{(k)}\!\left(f_\phi^{(k)}(x_c)
    \right) \right\|_2^2
\end{equation}

where $x_c$ denotes the operational feature vector of any record passed to autoencoder $k$. The anomaly threshold $\tau_k$ is set at the $p$-th percentile of reconstruction errors computed on $X_c^{(k)}$ \citep{Sakurada2014}; $p = 99$ was chosen to ensure a conservative threshold, minimising the risk of discarding operationally valid synthetic records:

\begin{equation}
    \label{eq:ae_threshold}
    \tau_k = \mathrm{P}_{p}\!\left(\left\{e^{(k)}\!\left(x_c^{(i)}\right)
        \right\}_{x_c^{(i)} \in X_c^{(k)}}\right)
\end{equation}

Once all autoencoders are trained, the operational cleaning phase begins. A synthetic record $\tilde{x}$ from $\tilde{\mathcal{S}}^{\mathrm{valid}}_k$ is discarded as operationally invalid if $e^{(k)}(\tilde{x}_c) > \tau_k$, where $\tilde{x}_c$ is its operational feature vector. This threshold retains the vast majority of operationally plausible synthetic records while removing those that no physically consistent flight could produce. Applying this cleaning step to each $\tilde{\mathcal{S}}^{\mathrm{valid}}_k$ independently yields the corresponding cleaned datasets $\tilde{\mathcal{S}}^{\mathrm{cleaned}}_k$ for $k = 0,\ldots,N_{\mathrm{tf}}$, which are subsequently used to construct the output datasets described in Section~\ref{sec:output}.

\subsubsection{Output}
\label{sec:output}
The pipeline produces three datasets used to evaluate the usefulness of the \emph{TailBooster} framework:

\smallskip
\noindent\textbf{(1) Na\"{i}ve Synthetic ($\mathcal{S}_{\text{na\"{i}ve}}$).} As defined in Section~\ref{sec:validity_filter}, $\mathcal{S}_{\text{na\"{i}ve}} = \tilde{\mathcal{S}}^{\mathrm{valid}}_0$ is the validity-filtered output of $\mathcal{G}_0$, the generative model trained on the full historical record $\mathcal{D}_0$, without operational cleaning or targeted extreme value augmentation. The Na\"{i}ve Synthetic dataset serves as the baseline representative of conventional synthetic data generation, against which the improvements introduced by the full \emph{TailBooster} pipeline are assessed.

\smallskip
\noindent\textbf{(2) Augmented Synthetic ($\mathcal{S}_{\mathrm{aug}}$).} Synthetic records $\tilde{\mathcal{S}}^{\mathrm{cleaned}}_k$ for $k = 0,\ldots,N_{\mathrm{tf}}$ that survive both the validity filter and the second anomaly detection layer are merged to constitute the Augmented Synthetic dataset, denoted $\mathcal{S}_{\mathrm{aug}}$. This dataset is the primary output of \emph{TailBooster}: by comparing it against $\mathcal{S}_{\text{na\"{i}ve}}$, the evaluation isolates the improvement that the dual-layer framework achieves over conventional synthetic data generation.

\smallskip
\noindent\textbf{(3) Augmented Real ($\mathcal{D}_{\mathrm{aug}}$).} The cleaned synthetic extreme records $\tilde{\mathcal{S}}^{\mathrm{cleaned}}_k$ for $k = 1,\ldots,N_{\mathrm{tf}}$, produced by the generative models trained on each $\mathcal{E}^{(k)}$ and surviving both filtering stages, are combined with the real historical data to form the Augmented Real dataset, denoted $\mathcal{D}_{\mathrm{aug}}$. This dataset demonstrates the utility gain achievable when the real historical record is directly enriched with operationally valid synthetic extremes produced by \emph{TailBooster}.

In addition to these three pipeline outputs, the original Real dataset $\mathcal{D}$ serves as the reference for all comparative evaluations. Table~\ref{tab:data_size} summarises the size of each of the four datasets and the number of extreme records per target feature; they are subsequently compared across the evaluation dimensions described in Section~\ref{sec:evaluation} to assess the contribution of extreme augmentation and operational cleaning to operational validity and downstream predictive performance on extreme values.

\begin{table}[H]
  \centering
  \footnotesize
  \resizebox{\columnwidth}{!}{%
  \begin{tabular}{@{}lccc@{}}
    \toprule
    \textbf{Dataset} &
    \textbf{Records} &
    \makecell[c]{\textbf{Extremes of} \\ \textbf{``Air Time (min)''}} &
    \makecell[c]{\textbf{Extremes of} \\ \textbf{``Arrival $\Delta$T (min)''}} \\
    \midrule
    Real ($\mathcal{D}$)                                    & \num{60767} & \num{3726} & \num{5470}  \\
    Na\"{i}ve synthetic ($\mathcal{S}_{\text{na\"{i}ve}}$)  & \num{62803} & \num{2751} & \num{5136}  \\
    Augmented synthetic ($\mathcal{S}_{\mathrm{aug}}$)      & \num{68335} & \num{6255} & \num{9938}  \\
    Augmented real ($\mathcal{D}_{\mathrm{aug}}$)           & \num{69390} & \num{7509} & \num{10546} \\
    \bottomrule
  \end{tabular}}
  \caption{Size of each dataset and number of extreme records per
  target feature. All datasets comprise 30 features.}
  \label{tab:data_size}
\end{table}

\begin{algorithm}[H]
\caption{\emph{TailBooster}}
\label{alg:tailbooster}
\begin{algorithmic}[1]

\Statex \hspace{-\algorithmicindent}\textbf{Input:}
\Statex $\mathcal{D} = \{x^{(i)}\}_{i=1}^{N}$ — Real dataset
\Statex $\mathcal{T} = \{f_1, \ldots, f_{N_{\mathrm{tf}}}\}$ — target features
\Statex $\mathcal{X}_c$ — operationally correlated feature names
\Statex $p = 99$ — anomaly threshold percentile
\Statex $r = 1.2$ — sampling ratio
\Statex \hspace{-\algorithmicindent}\textbf{Output:}
\Statex $\mathcal{S}_{\text{na\"{i}ve}}$,\;
             $\mathcal{S}_{\mathrm{aug}}$,\; $\mathcal{D}_{\mathrm{aug}}$

\vspace{4pt}
\Statex \hspace{-\algorithmicindent}\textbf{IQR-based extreme subset extraction}
       \textit{(1st anomaly detection layer)}
\For{each target feature $f_j \in \mathcal{T}$}
    \State Compute $Q_1^{(j)}$, $Q_3^{(j)}$, and
           $\mathrm{IQR}^{(j)} \gets Q_3^{(j)} - Q_1^{(j)}$
    \State Extract $\mathcal{E}^{(j)}$ using
           Equation~\eqref{eq:extreme_subset}
\EndFor

\State Form collection
       $\mathbb{D} \gets \{\mathcal{D}_0,\,\mathcal{E}^{(1)},\,\ldots,\,
       \mathcal{E}^{(N_{\mathrm{tf}})}\}$
       where $\mathcal{D}_0 = \mathcal{D}$

\vspace{4pt}
\Statex \hspace{-\algorithmicindent}\textbf{Autoencoder training}
       \textit{(2nd anomaly detection layer --- training phase)}
\Statex \textit{// Autoencoder training precedes generative model training in the execution order, as each autoencoder is trained once on real data and then applied post-generation for operational cleaning.}

\For{each dataset $\mathcal{D}_k \in \mathbb{D}$}
    \State Extract matrix $X_c^{(k)}$ from $\mathcal{D}_k$
           using feature names $\mathcal{X}_c$
    \State Train autoencoder $(f_\phi^{(k)},\,g_\theta^{(k)})$
           on $X_c^{(k)}$ by minimising
           $\mathcal{L}_{\mathrm{AE}}$
           (Equation~\eqref{eq:ae_loss})
    \State Compute $\tau_k \gets \mathrm{P}_{p}
           \bigl(\{e^{(k)}(x_c^{(i)})\}_{x_c^{(i)}\in X_c^{(k)}}\bigr)$
           using Equation~\eqref{eq:ae_threshold}
\EndFor

\vspace{4pt}
\Statex \hspace{-\algorithmicindent}\textbf{Generative model training} \textit{(default: TVAE)}
\For{each dataset $\mathcal{D}_k \in \mathbb{D}$}
    \State Train generative model $\mathcal{G}_k$ on $\mathcal{D}_k$
    \State Set sample size $n_k \gets \lfloor r \cdot |\mathcal{D}_k| \rfloor$
    \State Sample synthetic dataset
           $\tilde{\mathcal{S}}_k \gets \mathcal{G}_k(n_k)$
\EndFor
\Statex \textit{// $k{=}0$: $\tilde{\mathcal{S}}_0$ is sampled from $\mathcal{G}_0$ trained on the full historical record $\mathcal{D}_0 = \mathcal{D}$, with extremes of $\mathcal{T}$ under-represented}
\Statex \textit{// $k = 1,\ldots,N_{\mathrm{tf}}$: $\tilde{\mathcal{S}}_k$ are synthetic extremes corresponding to $\mathcal{E}^{(k)}$}

\vspace{4pt}
\Statex \hspace{-\algorithmicindent}\textbf{Relational validity filtering and operational cleaning}
       \textit{(2nd anomaly detection layer --- inference phase)}
\State Initialise $\mathcal{S}_{\mathrm{aug}} \gets \emptyset$
\For{each $\tilde{\mathcal{S}}_k \in
       \{\tilde{\mathcal{S}}_0,\,\tilde{\mathcal{S}}_1,\,\ldots,\,
       \tilde{\mathcal{S}}_{N_{\mathrm{tf}}}\}$}
    \Statex \textit{// Relational validity filter}
    \State $\tilde{\mathcal{S}}^{\mathrm{valid}}_k \gets
           \{\tilde{x} \in \tilde{\mathcal{S}}_k :
           (\tilde{o},\tilde{d}) \in \mathcal{P}_{\mathcal{D}}\}$
    \Statex \textit{// Operational cleaning}
    \State $\tilde{\mathcal{S}}^{\mathrm{cleaned}}_k \gets
           \{\tilde{x} \in \tilde{\mathcal{S}}^{\mathrm{valid}}_k :
           e^{(k)}(\tilde{x}_c) \leq \tau_k\}$
    \State $\mathcal{S}_{\mathrm{aug}} \gets
           \mathcal{S}_{\mathrm{aug}} \cup
           \tilde{\mathcal{S}}^{\mathrm{cleaned}}_k$
           \quad \textit{// ``augmented synthetic'' dataset}
\EndFor
\State $\mathcal{S}_{\text{na\"{i}ve}} \gets \tilde{\mathcal{S}}^{\mathrm{valid}}_0$
       \quad \textit{// ``na\"{i}ve synthetic'' dataset}

\State $\mathcal{D}_{\mathrm{aug}} \gets \mathcal{D} \cup
       \bigl(\bigcup_{k=1}^{N_{\mathrm{tf}}}
       \tilde{\mathcal{S}}^{\mathrm{cleaned}}_k\bigr)$
       \quad \textit{// ``augmented real'' dataset}

\vspace{4pt}
\State \Return $\mathcal{S}_{\text{na\"{i}ve}}$,\;
               $\mathcal{S}_{\mathrm{aug}}$,\;
               $\mathcal{D}_{\mathrm{aug}}$
\end{algorithmic}
\end{algorithm}

\renewcommand{\thesubsubsection}{\thesubsection.\arabic{subsubsection}}

\subsection{Evaluation Framework}
\label{sec:evaluation}
Credibility is a prerequisite for any synthetic dataset intended to support downstream analysis; without it, the generated records risk propagating structural errors into any model trained on them. Accordingly, each generated dataset was subjected to a set of structural integrity checks, verifying that the number of features was preserved, that continuous variables remained within the ranges observed in the real data, and that discrete attributes were confined to their original categories. Beyond these structural checks, five complementary dimensions of quality were assessed. As summarised in Table~\ref{tab:eval_framework}, these dimensions are not evaluated with the same objective: diversity, statistical similarity, and fidelity serve as preservation checks, confirming that the gains achieved in the target dimensions do not come at the cost of losing qualities that conventional generation already achieves, while operational validity and utility constitute the primary improvement targets of \emph{TailBooster}.

\begin{table}[H]
  \centering
  \footnotesize
  \resizebox{\columnwidth}{!}{%
  \begin{tabular}{@{}lc@{}}
    \toprule
    \textbf{Evaluation dimension} & \textbf{\emph{TailBooster} scope} \\
    \midrule
    Diversity (coverage of historical patterns)    & To preserve \\
    Statistical similarity                         & To preserve \\
    Fidelity (indistinguishability from real data) & To preserve \\
    \midrule
    Operational validity of synthetic records & To improve \\
    Utility in predicting extremes            & To improve \\
    \bottomrule
  \end{tabular}}
  \caption{Scope of {\normalfont\textit{TailBooster}} framework, relative to
  conventional generative approaches.}
  \label{tab:eval_framework}
\end{table}

\subsubsection{Diversity Assessment}
\label{sec:diversity_methodology}
Diversity constitutes the first of three preservation checks. A synthetic dataset is considered diverse if its generated records span the full range of patterns present in the real data. It was assessed through visual inspection of distributional coverage across three pairwise comparisons: (i)~Real vs.\ Na\"{i}ve Synthetic, (ii)~Real vs.\ Augmented Synthetic, and (iii)~Real vs.\ Augmented Real. In each comparison, two complementary dimensionality reduction methods were applied to both datasets: Principal Component Analysis (PCA) \citep{wold1987principal}, which recovers linear structure, and t-distributed Stochastic Neighbour Embedding (t-SNE) \citep{van2008visualizing}, which preserves non-linear neighbourhood relationships. Both methods were used to project the datasets into two-dimensional space, thereby allowing direct visual comparison of the underlying clusters. While improving diversity is not an objective of \emph{TailBooster}, this dimension was evaluated to confirm that the process of extremes augmentation and operational cleaning does not degrade the range of representation already present in the real data; limited variability in synthetic records risks introducing systematic bias into any downstream model trained on them.

\subsubsection{Statistical Assessment}
\label{sec:statistical_methodology} 
The second preservation check concerns statistical similarity: how closely the distributions of synthetic and real records align, assessed at both the marginal and bivariate levels. The same three pairwise comparisons were conducted: (i)~Real vs.\ Na\"{i}ve Synthetic, (ii)~Real vs.\ Augmented Synthetic, and (iii)~Real vs.\ Augmented Real. At the univariate level, the Kolmogorov--Smirnov test was applied to numerical and datetime features \citep{viehmann2021numerically}, while Total Variation Distance was used for categorical and boolean attributes \citep{Knoblauch2020RobustBI}. Pairwise relationships were further quantified through Correlation Similarity for numerical feature pairs \citep{sdmetrics_correlationsimilarity} and Contingency Similarity for categorical ones \citep{sdmetrics_contingencysimilarity}. Although \emph{TailBooster} does not target improvement in statistical similarity, this dimension was assessed to verify that the augmentation process does not distort the distributional characteristics of the real data; evaluating both univariate and bivariate agreement guarded against the misleading conclusions that can arise when distributional similarity is judged solely at the individual feature level.

\subsubsection{Fidelity Assessment}
\label{sec:fidelity_methodology}
The third and final preservation check is fidelity. A synthetic dataset exhibits high fidelity when its records are statistically indistinguishable from real ones; it was measured by training a binary classifier to discriminate between the two, on the basis that a classifier that cannot reliably distinguish between them is taken as evidence of higher fidelity in the generated data. A Random Forest classifier was employed given its well-established capacity to model complex, non-linear feature interactions \citep{breiman2001random}. To account for the size disparity between the compared datasets and to ensure robust estimates, stratified five-fold cross-validation with shuffling was applied \citep{Ahmadi2024ACS}. Performance was quantified using the F1 score and balanced accuracy \citep{sokolova2009systematic}; the latter assigns equal weight to both classes irrespective of their relative frequency, making it appropriate for imbalanced classification tasks. Two comparisons were evaluated: (i)~Real vs.\ Na\"{i}ve Synthetic, and (ii)~Real vs.\ Augmented Synthetic. The Real vs.\ Augmented Real comparison was omitted because real instances are shared between both datasets, rendering the classification task uninformative. Each comparison was repeated across three data subsets: the full dataset (nominal and extreme cases combined), records with extreme ``Air Time (min)'', and records with extreme ``Arrival $\Delta$T (min)'', yielding six tests in total.

Indistinguishability alone does not preclude memorisation, since a generative model could achieve high fidelity scores simply by reproducing training samples. To guard against this, the Distance to Closest Record (DCR) was computed for each synthetic sample \citep{park2018data} as a test of memorisation. DCR measures the Euclidean distance between each synthetic record and its closest counterpart in the real dataset, computed exclusively on numerical features after applying $z$-score normalisation fitted to the real data. Each cleaned synthetic dataset ($\tilde{\mathcal{S}}^{\mathrm{cleaned}}_0$, $\tilde{\mathcal{S}}^{\mathrm{cleaned}}_1$, and $\tilde{\mathcal{S}}^{\mathrm{cleaned}}_2$) was evaluated against its paired real dataset ($\mathcal{D}_0$, $\mathcal{E}^{(1)}$, and $\mathcal{E}^{(2)}$, respectively), which was used to train the corresponding generative model. To establish a reference, a real-to-real baseline mean $\bar{d}_{\mathrm{real},k}$ was computed as the mean nearest-neighbour distance among real records in each paired dataset, excluding self-matches \citep{platzer2021holdout}. The DCR ratio is defined as the mean synthetic-to-real DCR normalised by $\bar{d}_{\mathrm{real},k}$; a ratio greater than or equal to~1 indicates that synthetic samples are, on average, no closer to real records than real samples are to one another, providing evidence against memorisation. As a secondary check, a memorisation threshold $\tau_{\mathrm{DCR},k} = 0.10 \times \bar{d}_{\mathrm{real},k}$ was defined for each synthesiser $k \in \{0, 1, 2\}$; any synthetic record with DCR below $\tau_{\mathrm{DCR},k}$ was flagged as a potential near-copy and subsequently inspected manually to distinguish memorised copies from near-neighbours differing only in continuous time-valued fields. While neither fidelity improvement nor memorisation prevention is a primary objective of the proposed framework, measuring both serves as a safeguard against unintended quality degradation introduced by the augmentation and cleaning process.

\subsubsection{Operational Validity Assessment}
\label{sec:operational_methodology}
Operational validity is the first of \emph{TailBooster}'s two primary improvement targets. A synthetic flight record is considered operationally valid if it respects the empirical correlations between operational attributes observed in the real data, such as the expected relationship between flight distance and air time. This was evaluated by examining two pairwise operational correlations: ``Air Time (min)'' against ``Distance (miles)'', and ``Arrival $\Delta$T (min)'' against ``Distance (miles)'', across three comparisons: (i)~Real vs.\ Na\"{i}ve Synthetic, (ii)~Real vs.\ Augmented Synthetic, and (iii)~Real vs.\ Augmented Real. In our prior work \citep{Aly2025a, Aly2026}, synthetic datasets consistently contained implausible records, such as unrealistically short flight times for long-haul routes, indicating violations of operational feasibility. Addressing this limitation is a central objective of the \emph{TailBooster} framework, specifically the second anomaly detection layer, which enforces operational feasibility by filtering records that violate the empirical operational envelope learned from the historical data.

\subsubsection{Utility Assessment}
\label{sec:utility_methodology}
Utility in predicting extreme values is \emph{TailBooster}'s principal improvement target. Concretely, it measures the reduction in prediction error---quantified via Mean Absolute Error (MAE)---on extreme values achieved by training on augmented data, relative to baselines trained on real or na\"{i}ve synthetic data alone, reflecting the practical benefit of more accurate prediction of severe delays and air times. Both training and testing were conducted exclusively on the extremes of each target feature, as nominal records' numerical dominance would otherwise mask genuine gains in extreme-value prediction.

Utility was assessed by comparing four scenarios: (i)~Train on Real, Test on Real; (ii)~Train on Na\"{i}ve Synthetic, Test on Real; (iii)~Train on Augmented Synthetic, Test on Real; and (iv)~Train on Augmented Real, Test on Real. This four-scenario structure was applied once to records with extreme ``Air Time (min)'' and again to records with extreme ``Arrival $\Delta$T (min)''. Only the real dataset was partitioned into training and test subsets; all other datasets were used exclusively for training. Shuffling was applied during the real-data split to ensure that the test set captured the full spectrum of extreme values. Six regression algorithms were employed: (i)~Random Forest Regression \citep{breiman2001random}, which reduces variance through bagging and is robust to overfitting in mixed-type feature spaces; (ii)~XGBoost Regression \citep{Chen2016}, a regularised gradient-boosted ensemble with strong general-purpose predictive performance; (iii)~CatBoost Regression \citep{Prokhorenkova2018}, which natively handles categorical features without requiring manual encoding; (iv)~LightGBM Regression \citep{Ke2017}, selected for its computational efficiency and effective handling of mixed feature types; (v)~Support Vector Regression \citep{Drucker1996}, chosen for its robustness to outliers via the epsilon-insensitive loss formulation; and (vi)~k-Nearest Neighbours Regression \citep{CoverHart1967}, which provides a non-parametric, assumption-free baseline sensitive to local structural patterns. Examining utility across multiple regression algorithms ensures that the findings are not specific to any single model family and that the observed gains in extreme-value prediction are robust across a range of inductive biases.

Prediction was conducted in the tactical phase, using only information available at the time of take-off (Table~\ref{tab:features}); any time-related feature that could indirectly encode the actual arrival time was withheld to guard against information leakage. This constraint is consistent with the requirements of real-world operational deployment \citep{MASPUJOL2024124146}. The four-scenario comparison provides direct evidence of the contribution made by the operational cleaning process and the extreme-value augmentation to the prediction of operationally significant events, specifically extreme air times and extreme arrival delays.

\subsection{Hyperparameter Tuning}
\label{sec:optimisation}
The hyperparameters of each generative model were tuned using the multi-objective optimisation framework we proposed in \citet{Aly2026}, employing the Tree-structured Parzen Estimator (TPE) algorithm \citep{akiba2019optuna}. TPE constructs two non-parametric density estimators over the hyperparameter space: one for configurations associated with good performance, and another for those with poor performance. New trials are sampled from regions where the ratio between these densities is maximised, making TPE more sample-efficient than grid or random search and well suited to high-dimensional hyperparameter spaces with mixed variable types \citep{bergstra2011algorithms}. In \citet{Aly2026}, we provided a detailed justification of the objective function weights and a sensitivity analysis of synthetic data quality across different weight configurations.

Three TVAE hyperparameters were included in the search: the embedding dimension, the encoder layer widths (two layers, tuned independently), and the number of training epochs. The decoder layer widths were fixed as the mirror of the encoder, ensuring a symmetric architecture and reducing the search space without introducing an additional degree of freedom. Table~\ref{tab:search_space} summarises the search space for all three parameters.

\begin{table}[H]
  \centering
  \footnotesize
  \resizebox{\columnwidth}{!}{%
  \begin{tabular}{@{}ll@{}}
    \toprule
    \textbf{Hyperparameter} & \textbf{Search space} \\
    \midrule
    Embedding dimension        & $\{8, 16, 32, 64, 128, 200, 256, 300, 400, 500, 600\}$ \\
    Encoder layer width (each) & $\{8, 16, 32, 64, 128, 200, 256, 300, 400, 500, 600\}$ \\
    Training epochs            & $\{300, 400, \ldots, 6000\}$ (step: 100) \\
    \bottomrule
  \end{tabular}}
  \caption{TVAE hyperparameter search space.}
  \label{tab:search_space}
\end{table}

Each optimisation run was capped at 100~trials. The 10~initial trials were allocated to random exploration before TPE began directing the search, following the standard warm-up protocol \citep{akiba2019optuna}.

\section{Results}
\label{sec:results}
This section evaluates the \emph{TailBooster} framework by comparing the four datasets defined in Section~\ref{sec:output}: Real ($\mathcal{D}$), Na\"{i}ve Synthetic ($\mathcal{S}_{\text{na\"{i}ve}}$), Augmented Synthetic ($\mathcal{S}_{\mathrm{aug}}$), and Augmented Real ($\mathcal{D}_{\mathrm{aug}}$). The comparison between $\mathcal{S}_{\text{na\"{i}ve}}$ and $\mathcal{S}_{\mathrm{aug}}$ isolates the contribution of the proposed dual-layer pipeline over conventional synthetic data generation. Since both datasets pass through the relational validity filter, this comparison reflects the added benefit of both the extreme-value augmentation and operational cleaning. The comparison between $\mathcal{D}$ and $\mathcal{D}_{\mathrm{aug}}$ quantifies the benefit of enriching the real historical record with operationally valid synthetic extremes. The results are organised into five subsections, each corresponding to one of the evaluation dimensions described in Section~\ref{sec:evaluation}. As summarised in Table~\ref{tab:eval_framework}, diversity, statistical similarity, and fidelity are assessed as preservation checks, while operational validity and utility constitute the primary improvement targets of the framework.

\subsection{Diversity Assessment}
\label{sec:diversity_results}
\begin{figure}[H]
    \centering
    \begin{minipage}{0.23\textwidth} 
        \centering
        \includegraphics[width=\textwidth, trim=0mm 0mm 0mm 0mm, clip]{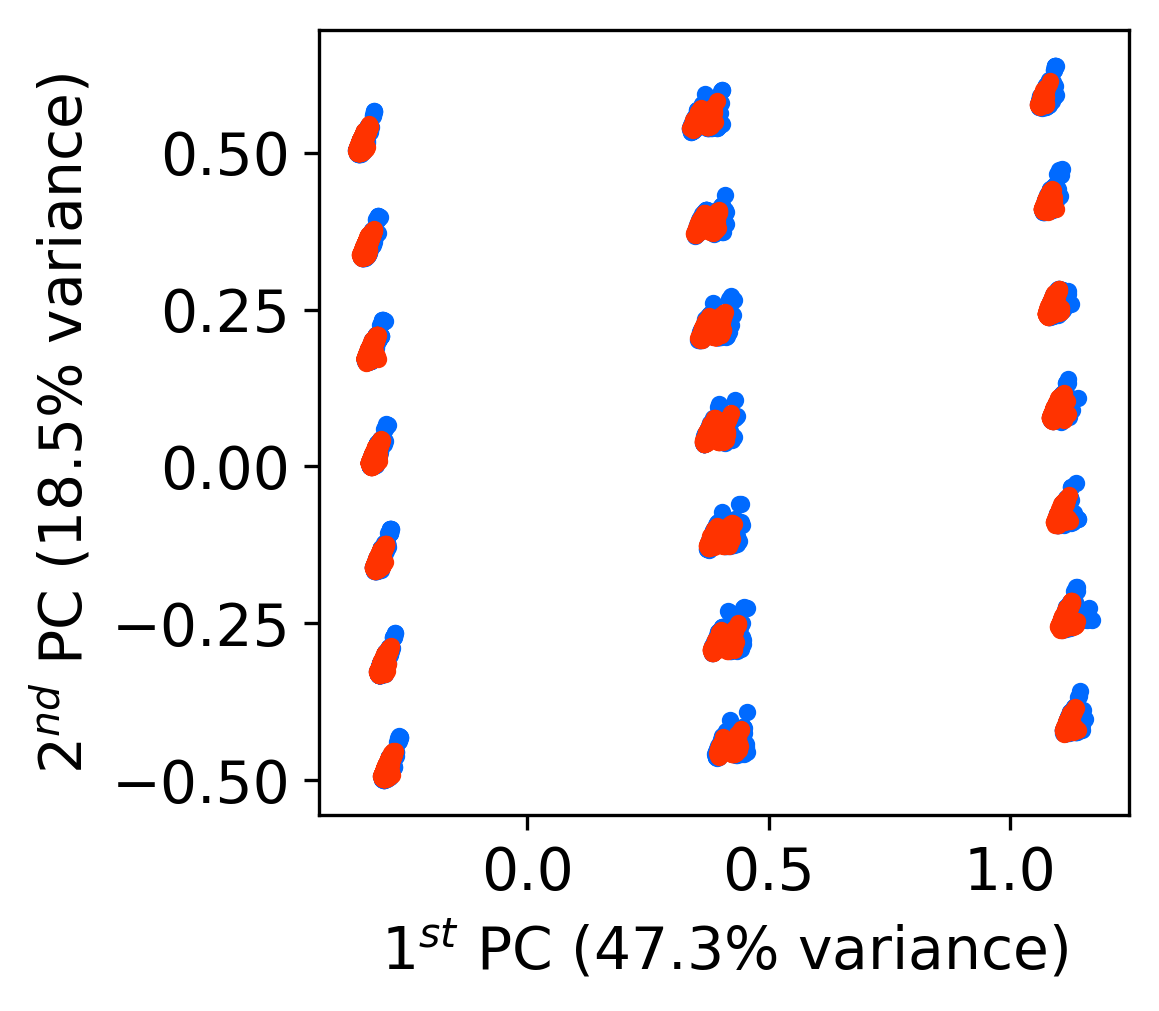} 
        \vspace{-0.7cm}
        \subcaption{\centering PCA: Real vs.\ Na\"{i}ve Synth.}
        \label{fig:diversity_1}
    \end{minipage}
    \hfill
    \begin{minipage}{0.23\textwidth} 
        \centering
        \includegraphics[width=\textwidth, trim=0mm 0mm 0mm 0mm, clip]{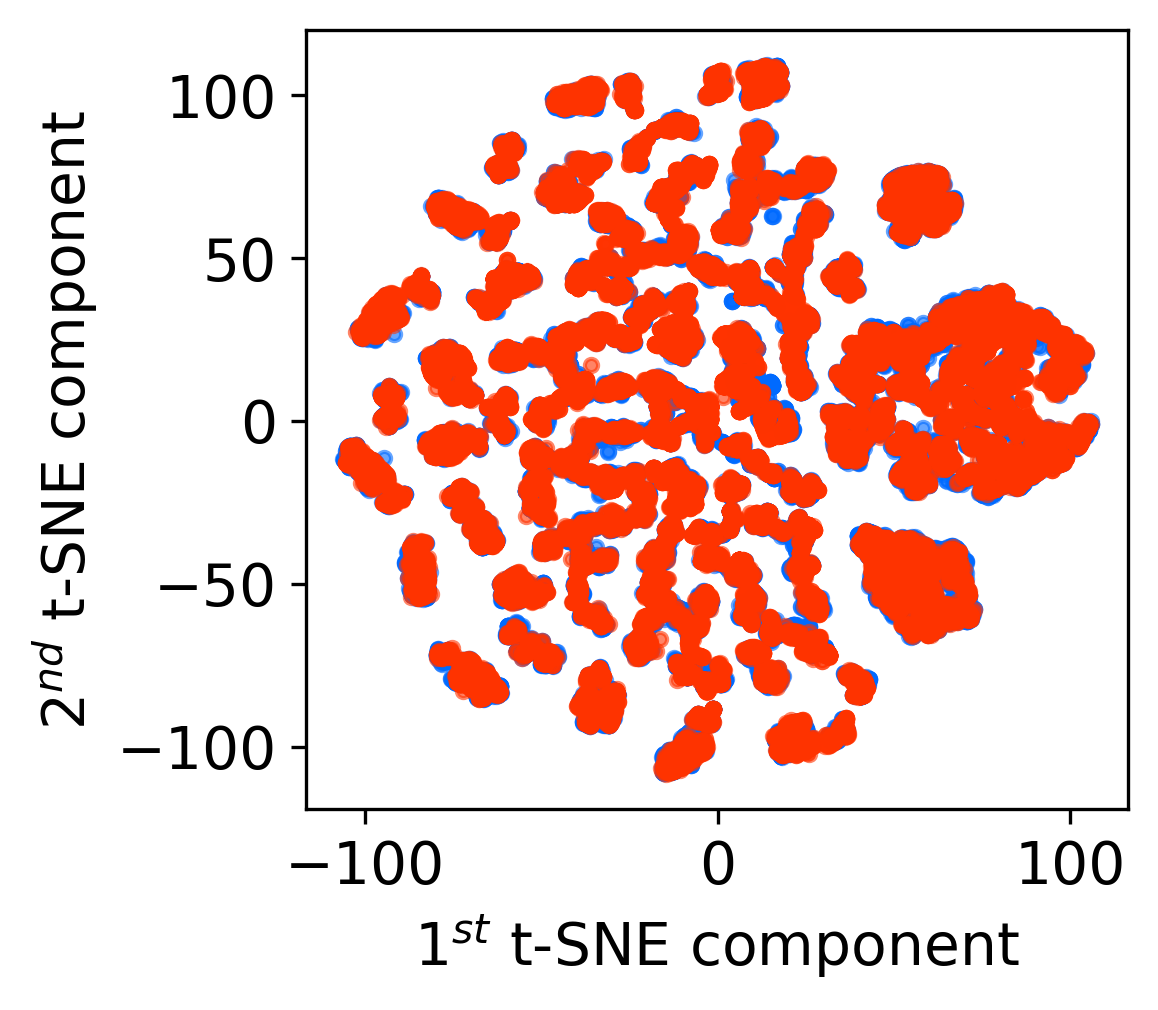} 
        \vspace{-0.7cm}
        \subcaption{\centering t-SNE: Real vs.\ Na\"{i}ve Synth.}
        \label{fig:diversity_2}
    \end{minipage}
    \hfill
    \begin{minipage}{0.23\textwidth} 
        \centering
        \includegraphics[width=\textwidth, trim=0mm 0mm 0mm 0mm, clip]{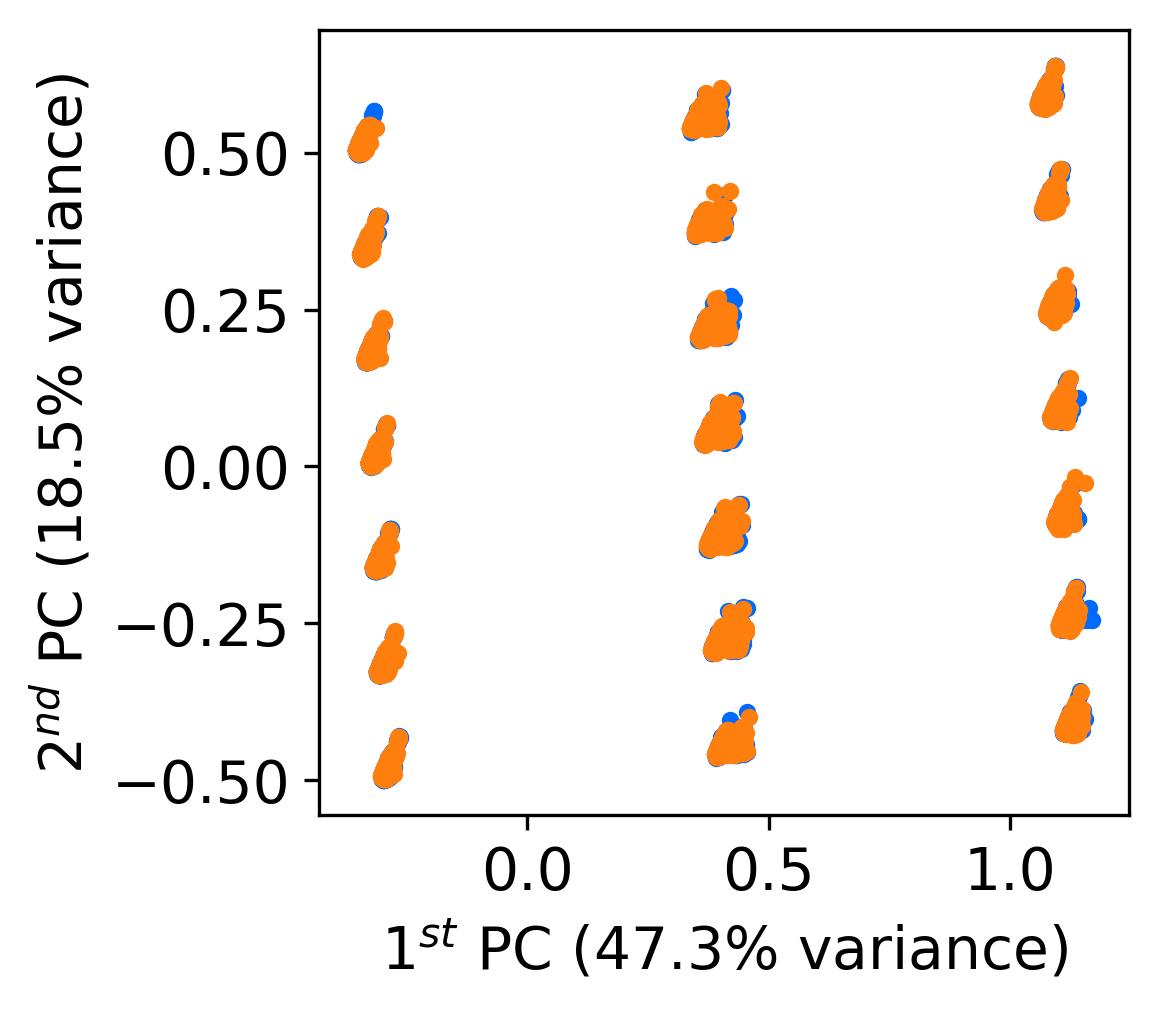} 
        \vspace{-0.7cm}
        \subcaption{\centering PCA: Real vs.\ Aug. Synth.}
        \label{fig:diversity_3}
    \end{minipage}
    \hfill 
    \begin{minipage}{0.23\textwidth} 
        \centering
        \includegraphics[width=\textwidth, trim=0mm 0mm 0mm 0mm, clip]{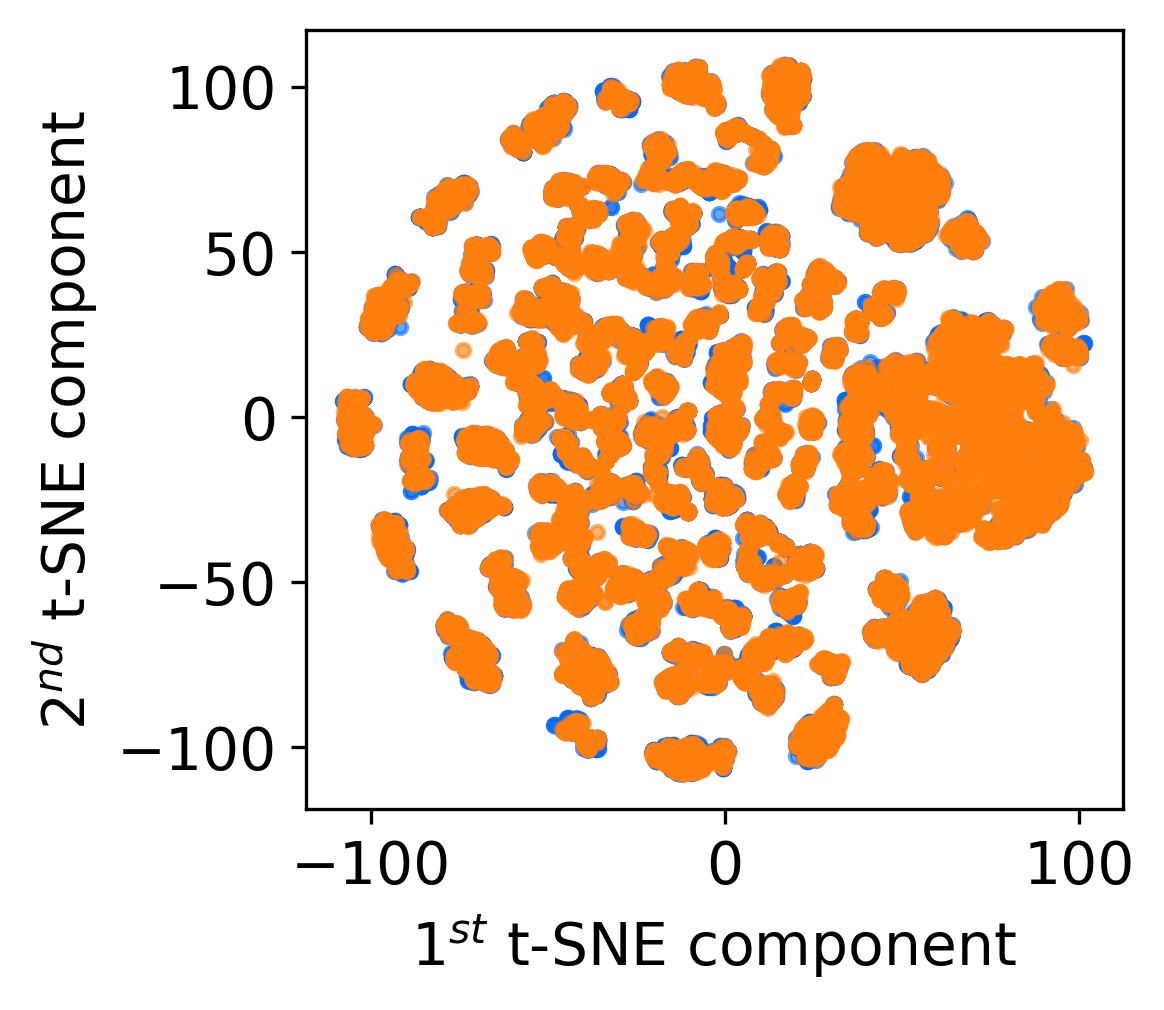} 
        \vspace{-0.7cm}
        \subcaption{\centering t-SNE: Real vs.\ Aug. Synth.}
        \label{fig:diversity_4}
    \end{minipage}
    \hfill
    \begin{minipage}{0.23\textwidth} 
        \centering
        \includegraphics[width=\textwidth, trim=0mm 0mm 0mm 0mm, clip]{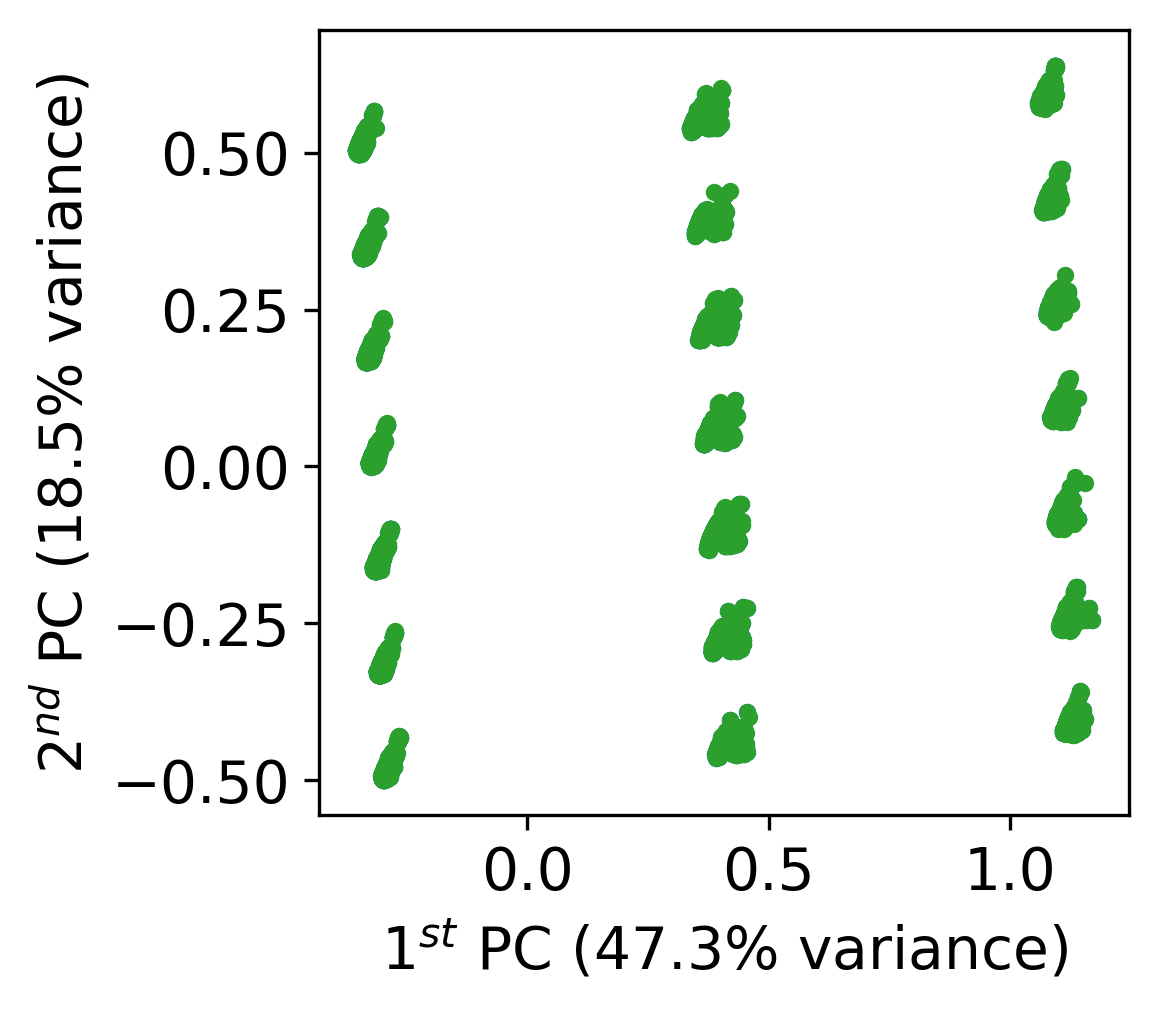} 
        \vspace{-0.7cm}
        \subcaption{\centering PCA: Real vs.\ Aug. Real}
        \label{fig:diversity_5}
    \end{minipage}
    \hfill 
    \begin{minipage}{0.23\textwidth} 
        \centering
        \includegraphics[width=\textwidth, trim=0mm 0mm 0mm 0mm, clip]{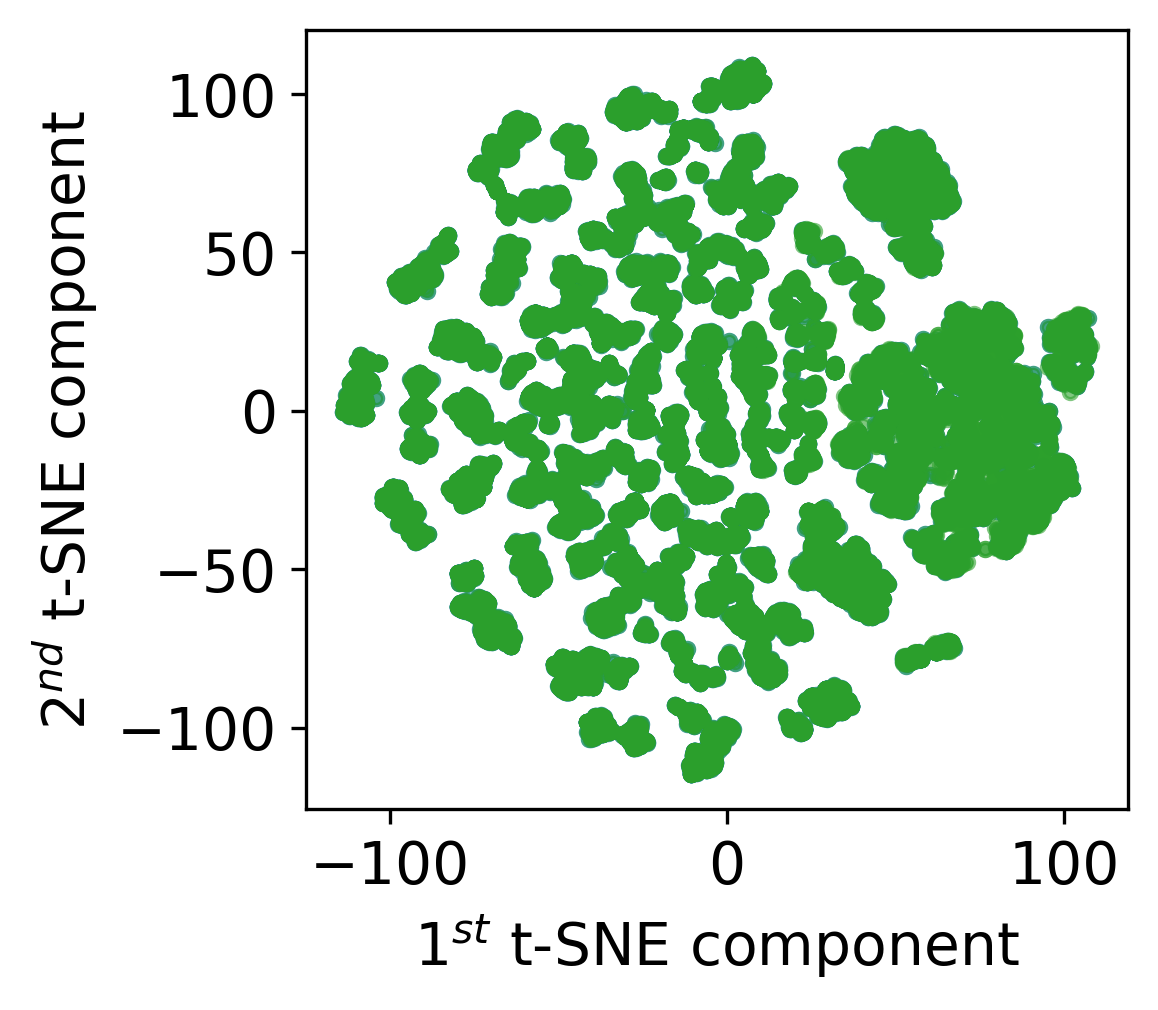} 
        \vspace{-0.7cm}
        \subcaption{\centering t-SNE: Real vs.\ Aug. Real}
        \label{fig:diversity_6}
    \end{minipage}
    \caption{\centering Diversity comparison of real (blue), na\"{i}ve synthetic (red), augmented synthetic (orange), and augmented real (green) flight records.}
    \label{fig:diversity}
\end{figure}

Figure~\ref{fig:diversity} presents the PCA and t-SNE projections for the three pairwise comparisons. In all panels, real data are plotted in blue beneath the comparison dataset, so that the degree to which the synthetic or augmented records cover the real clusters can be assessed visually.

The PCA projections reveal a clear discrete cluster structure, with three primary groupings along the first principal component, each internally stratified into several sub-clusters. In the Na\"{i}ve Synthetic comparison (Figure~\ref{fig:diversity_1}), synthetic records overlap with all real clusters, confirming that conventional generation---a single model trained on the full data, validity-filtered but with no extreme-value augmentation or operational cleaning---captures the broad structure of the feature space. The Augmented Synthetic comparison (Figure~\ref{fig:diversity_3}) shows improved within-cluster coverage: for instance, six of the seven sub-clusters in the left-hand grouping are fully covered, with the real data points no longer visibly exposed. The Augmented Real comparison (Figure~\ref{fig:diversity_5}) achieves complete coverage across all clusters by construction, as it incorporates the real historical records directly, with extremes better represented through augmentation. The corresponding t-SNE projections (Figures~\ref{fig:diversity_2},~\ref{fig:diversity_4}, and~\ref{fig:diversity_6}) confirm this pattern at the level of non-linear neighbourhood structure: all three datasets reproduce the broad topology of the real data manifold. Although diversity improvement was not an objective of the framework, these results confirm that the \emph{TailBooster} pipeline does not degrade diversity, with the Augmented Synthetic showing modest gains over the Na\"{i}ve Synthetic.

\subsection{Statistical Assessment}
\label{sec:statistical_results}
Table~\ref{tab:statistical} reports the statistical similarity scores across the same three pairwise comparisons. The second column reports marginal similarity, the mean of the Kolmogorov--Smirnov and Total Variation Distance scores across all features, measuring univariate distributional agreement. The third column reports bivariate similarity, the mean of Correlation Similarity and Contingency Similarity scores across all feature pairs, capturing pairwise inter-feature relationships. The fourth column reports overall similarity, the mean of the two.

\begin{table}[H]
  \centering
  \footnotesize
  \resizebox{\columnwidth}{!}{%
  \begin{tabular}{@{}lccc@{}}
    \toprule
    \textbf{Comparison} &
    \makecell[c]{\textbf{Marginal} \\ \textbf{Similarity (\%)}} &
    \makecell[c]{\textbf{Bivariate} \\ \textbf{Similarity (\%)}} &
    \makecell[c]{\textbf{Overall} \\ \textbf{Similarity (\%)}} \\
    \midrule
    Real vs.\ Na\"{i}ve Synthetic     & 86.43 & 73.53 & 79.98 \\
    Real vs.\ Augmented Synthetic     & 90.05 & 82.47 & 86.26 \\
    Real vs.\ Augmented Real          & 96.72 & 92.47 & 94.60 \\
    \bottomrule
  \end{tabular}}
  \caption{Statistical similarity scores across the three pairwise comparisons (higher is better). Overall similarity is the mean of marginal and bivariate similarity.}
  \label{tab:statistical}
\end{table}

All three comparisons achieved high statistical similarity scores, confirming that neither the augmentation nor the cleaning process corrupts the distributional characteristics of the real data. The Na\"{i}ve Synthetic dataset achieved an overall similarity of 79.98\%, reflecting the baseline performance of conventional generation. The Augmented Synthetic improved on this, reaching 86.26\%, suggesting that the dual-layer pipeline does not degrade and in fact marginally improves statistical similarity relative to the na\"{i}ve baseline. The Augmented Real achieved the highest overall similarity of 94.60\%, which is expected given that it directly incorporates the real historical records. The most notable gains across all three datasets are observed in bivariate similarity, where the Augmented Synthetic improved by 8.94 percentage points over the Na\"{i}ve Synthetic (82.47\% vs.\ 73.53\%), indicating that the operational cleaning process, by enforcing empirical inter-feature correlations, also strengthens the bivariate distributional agreement between synthetic and real records.

\subsection{Fidelity Assessment}
\label{sec:fidelity_results}
The discriminability check was conducted across six tests, as described in Section~\ref{sec:fidelity_methodology}, and the results are summarised in Table~\ref{tab:fidelity}. Understanding these results requires first considering the composition of each synthetic dataset. Recall from Figure~\ref{fig:architecture} that the na\"{i}ve synthetic dataset $\mathcal{S}_{\text{na\"{i}ve}} = \tilde{\mathcal{S}}^{\mathrm{valid}}_0$ is the validity-filtered output of $\mathcal{G}_0$, the generative model trained on the full historical dataset $\mathcal{D}$, with no operational cleaning or extreme-value augmentation; it represents the baseline output of conventional synthetic data generation. The augmented synthetic dataset $\mathcal{S}_{\mathrm{aug}}$, by contrast, merges the operationally cleaned outputs of three generative models: $\mathcal{G}_0$, trained on the full data, and $\mathcal{G}_1$ and $\mathcal{G}_2$, each trained exclusively on the real extreme subset of one target feature. Since both datasets pass through the relational validity filter, the fidelity difference between them reflects the effect of operational cleaning and extreme-value augmentation alone. Two distinct patterns emerged depending on whether the classifier operated on the full dataset---comprising nominal and extreme cases---or on the extreme subsets of each target feature.

\begin{table}[H]
  \centering
  \footnotesize
  \resizebox{\columnwidth}{!}{%
  \begin{tabular}{@{}llccc@{}}
    \toprule
    \textbf{Subset} & \textbf{Comparison} & \textbf{F1} &
    \makecell[c]{\textbf{Balanced} \\ \textbf{Accuracy}} &
    \makecell[c]{\textbf{Overall} \\ \textbf{Discriminability}} \\
    \midrule
    \multirow{2}{*}{\makecell[l]{Full dataset \\ (nominal + extremes)}}
      & Real vs.\ Na\"{i}ve Synthetic     & 0.88 & 0.88 & 0.88 \\
      & Real vs.\ Augmented Synthetic     & 0.77 & 0.79 & 0.78 \\
    \midrule
    \multirow{2}{*}{\makecell[l]{Extremes of \\ ``Air Time (min)''}}
      & Real vs.\ Na\"{i}ve Synthetic     & 0.94 & 0.89 & 0.92 \\
      & Real vs.\ Augmented Synthetic     & 0.47 & 0.62 & 0.54 \\
    \midrule
    \multirow{2}{*}{\makecell[l]{Extremes of \\ ``Arrival $\Delta$T (min)''}}
      & Real vs.\ Na\"{i}ve Synthetic     & 0.88 & 0.88 & 0.88 \\
      & Real vs.\ Augmented Synthetic     & 0.50 & 0.65 & 0.58 \\
    \bottomrule
  \end{tabular}}
  \caption{Discriminability  check results across all six tests (lower is better).
  Overall discriminability is the mean of F1 and balanced accuracy.}
  \label{tab:fidelity}
\end{table}

When evaluated on the full dataset, discriminability was high in both comparisons, with overall discriminability scores of 0.88 for Real vs.\ $\mathcal{S}_{\text{na\"{i}ve}}$ and 0.78 for Real vs.\ $\mathcal{S}_{\mathrm{aug}}$. This is expected: both synthetic datasets are dominated by nominal records generated by the same model $\mathcal{G}_0$, so the classifier faced a largely similar task in both cases. The modest reduction in discriminability when moving from $\mathcal{S}_{\text{na\"{i}ve}}$ to $\mathcal{S}_{\mathrm{aug}}$ reflects two compounding effects. First, out of \num{62803} records in $\mathcal{S}_{\text{na\"{i}ve}}$, \num{3091} were flagged as anomalies by the autoencoder-based cleaning layer for violating the operational correlations learned from the historical data and subsequently removed. Second, $\tilde{\mathcal{S}}^{\mathrm{cleaned}}_1$ and $\tilde{\mathcal{S}}^{\mathrm{cleaned}}_2$ contributed \num{3688} extreme ``Air Time (min)'' records and \num{4935} extreme ``Arrival $\Delta$T (min)'' records, respectively, both of higher fidelity and operational validity than the corresponding extreme records in $\mathcal{S}_{\text{na\"{i}ve}}$. Together, these changes make $\mathcal{S}_{\mathrm{aug}}$ harder to distinguish from the real data than $\mathcal{S}_{\text{na\"{i}ve}}$. However, since the extreme records from $\mathcal{G}_1$ and $\mathcal{G}_2$ constitute a minority of the full dataset, their contribution to discriminability reduction is diluted by the dominant nominal majority, which is similarly generated in both cases, limiting the overall gain.

The second pattern was observed when the evaluation was restricted to the extreme subsets. For the extremes of ``Air Time (min)'', overall discriminability fell from 0.92 for Real vs.\ $\mathcal{S}_{\text{na\"{i}ve}}$ to 0.54 for Real vs.\ $\mathcal{S}_{\mathrm{aug}}$; for the extremes of ``Arrival $\Delta$T (min)'', it fell from 0.88 to 0.58. These drops are substantially larger than those observed at the full-dataset level and reflect a fundamental difference in the origin of the extreme records being compared. In the na\"{i}ve case, the extreme subset of $\mathcal{S}_{\text{na\"{i}ve}}$ consists of records generated by $\mathcal{G}_0$ that happen to fall in the tail of the target feature distribution; since $\mathcal{G}_0$ was trained on the full data and never received tail-concentrated training signal, and since these records were not subject to operational cleaning, they are poor representatives of real extremes and are easily distinguished from them by the classifier. In the augmented case, the extreme subset of $\mathcal{S}_{\mathrm{aug}}$ mostly consists of records from $\mathcal{G}_1$ and $\mathcal{G}_2$, each trained directly on the corresponding real extreme subset and subsequently filtered by the relational validity filter and the autoencoder-based operational cleaning layer. These records faithfully capture the distributional characteristics of real extremes, making the classification task substantially harder and reducing discriminability accordingly. Although indistinguishability was not an optimisation target, the consistently lower discriminability of $\mathcal{S}_{\mathrm{aug}}$ relative to $\mathcal{S}_{\text{na\"{i}ve}}$, at both the full-dataset level and across both extreme subsets, confirms that \emph{TailBooster}'s targeted extreme augmentation yields fidelity gains beyond those achievable with conventional generation alone.

As noted in Section~\ref{sec:fidelity_methodology}, high indistinguishability alone does not rule out memorisation. Table~\ref{tab:dcr} reports the DCR-based memorisation check results for the three cleaned synthetic datasets. All three datasets produced DCR ratios greater than 1 (1.22, 1.28, and 1.15 for $\tilde{\mathcal{S}}^{\mathrm{cleaned}}_0$, $\tilde{\mathcal{S}}^{\mathrm{cleaned}}_1$, and $\tilde{\mathcal{S}}^{\mathrm{cleaned}}_2$, respectively), indicating that synthetic records were, on average, more dispersed from the real data manifold than real records were from one another. For $\tilde{\mathcal{S}}^{\mathrm{cleaned}}_1$ and $\tilde{\mathcal{S}}^{\mathrm{cleaned}}_2$, no record fell below the memorisation threshold $\tau_{\mathrm{DCR},k}$, providing no evidence of memorisation in the synthetic extremes. For $\tilde{\mathcal{S}}^{\mathrm{cleaned}}_0$, 14 records out of \num{59712} (0.02\%) were flagged as potential copies of real data. However, manual inspection revealed that 13 of these were flagged primarily due to an exact match in categorical and route attributes with their nearest real neighbour, while differing in continuous time-valued fields by no more than a few minutes; this pattern is consistent with generalisation rather than memorisation. Only one of the 14 flagged records was found to be an exact duplicate, corresponding to a rate of 0.002\% of the dataset; this isolated case is attributable to the high frequency of the corresponding origin--destination pair in the training data. Overall, these results indicate that the \emph{TailBooster} pipeline does not memorise its training data to any meaningful degree.

\begin{table}[H]
  \centering
  \footnotesize
  \resizebox{\columnwidth}{!}{%
  \begin{tabular}{@{}lcccc@{}}
    \toprule
    \textbf{Synthetic dataset} & 
    \textbf{Mean DCR} & 
    \textbf{DCR ratio} & 
    $\boldsymbol{\tau_{\mathrm{DCR},k}}$ & 
    \textbf{Flagged (\%)} \\
    \midrule
    $\tilde{\mathcal{S}}^{\mathrm{cleaned}}_0$    & 0.5191 & 1.2212 & 0.0425 & 0.02 \\
    $\tilde{\mathcal{S}}^{\mathrm{cleaned}}_1$    & 0.8624 & 1.2763 & 0.0676 & 0.00 \\
    $\tilde{\mathcal{S}}^{\mathrm{cleaned}}_2$    & 0.6310 & 1.1470 & 0.0550 & 0.00 \\
    \bottomrule
  \end{tabular}}
  \caption{DCR-based memorisation check per synthetic dataset. Each dataset is evaluated against its paired real training set. DCR ratio $\geq 1$ indicates no systematic proximity to real records. Records with DCR $< \tau_{\mathrm{DCR},k}$ are flagged as potential near-copies.}
  \label{tab:dcr}
\end{table}

\subsection{Operational Assessment}
\label{sec:operational_results}

\begin{figure}[H]
    \centering
    \begin{minipage}{0.23\textwidth} 
        \centering
        \includegraphics[width=\textwidth, trim=0mm 0mm 0mm 0mm, clip]{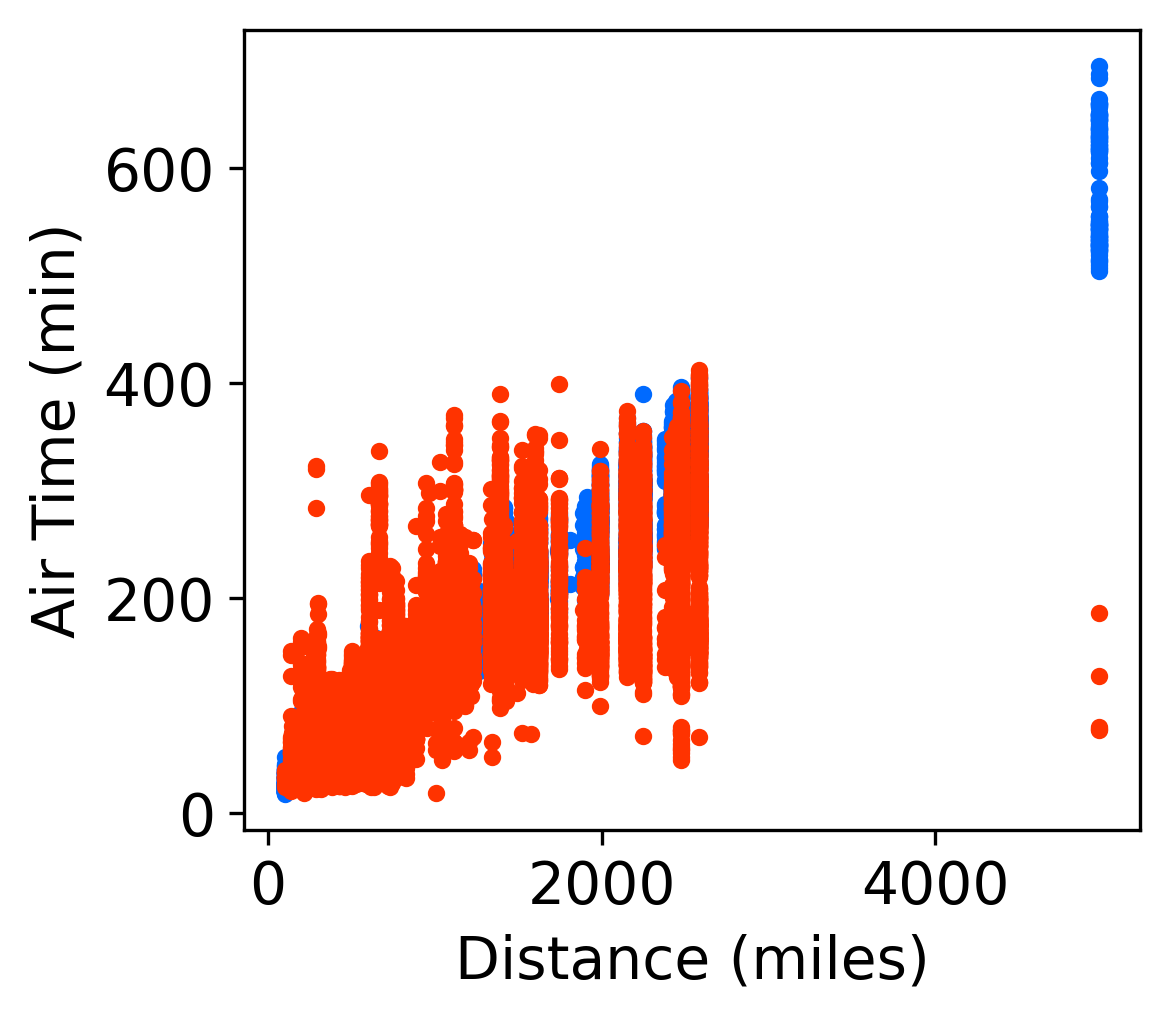} 
        \vspace{-0.7cm}
        \subcaption{\centering Real vs.\ Na\"{i}ve Synth.}
        \label{fig:operational_1}
    \end{minipage}
    \hfill
    \begin{minipage}{0.23\textwidth} 
        \centering
        \includegraphics[width=\textwidth, trim=0mm 0mm 0mm 0mm, clip]{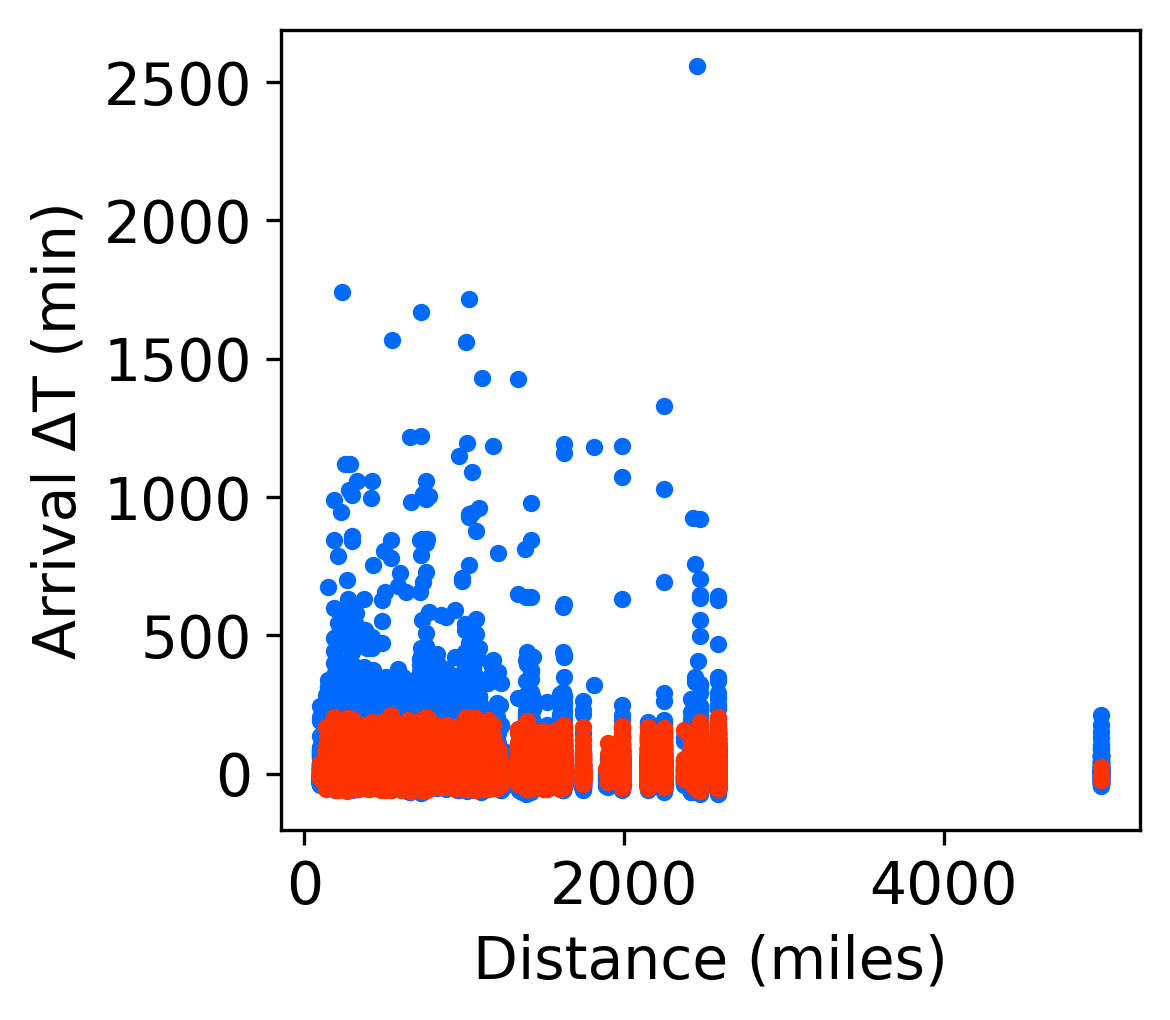} 
        \vspace{-0.7cm}
        \subcaption{\centering Real vs.\ Na\"{i}ve Synth.}
        \label{fig:operational_2}
    \end{minipage}
    \hfill
    \begin{minipage}{0.23\textwidth} 
        \centering
        \includegraphics[width=\textwidth, trim=0mm 0mm 0mm 0mm, clip]{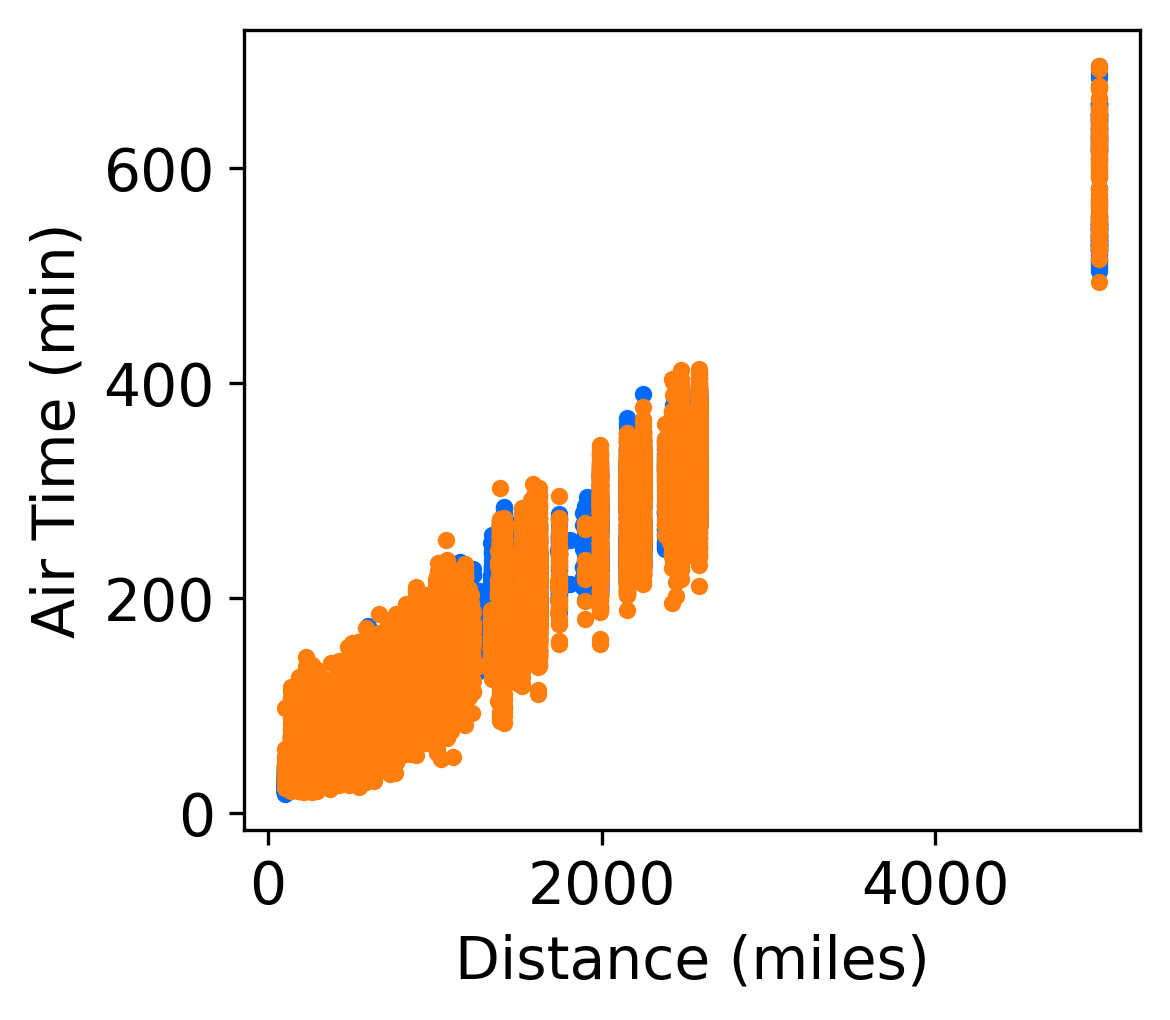} 
        \vspace{-0.7cm}
        \subcaption{\centering Real vs.\ Aug. Synth.}
        \label{fig:operational_3}
    \end{minipage}
    \hfill 
    \begin{minipage}{0.23\textwidth} 
        \centering
        \includegraphics[width=\textwidth, trim=0mm 0mm 0mm 0mm, clip]{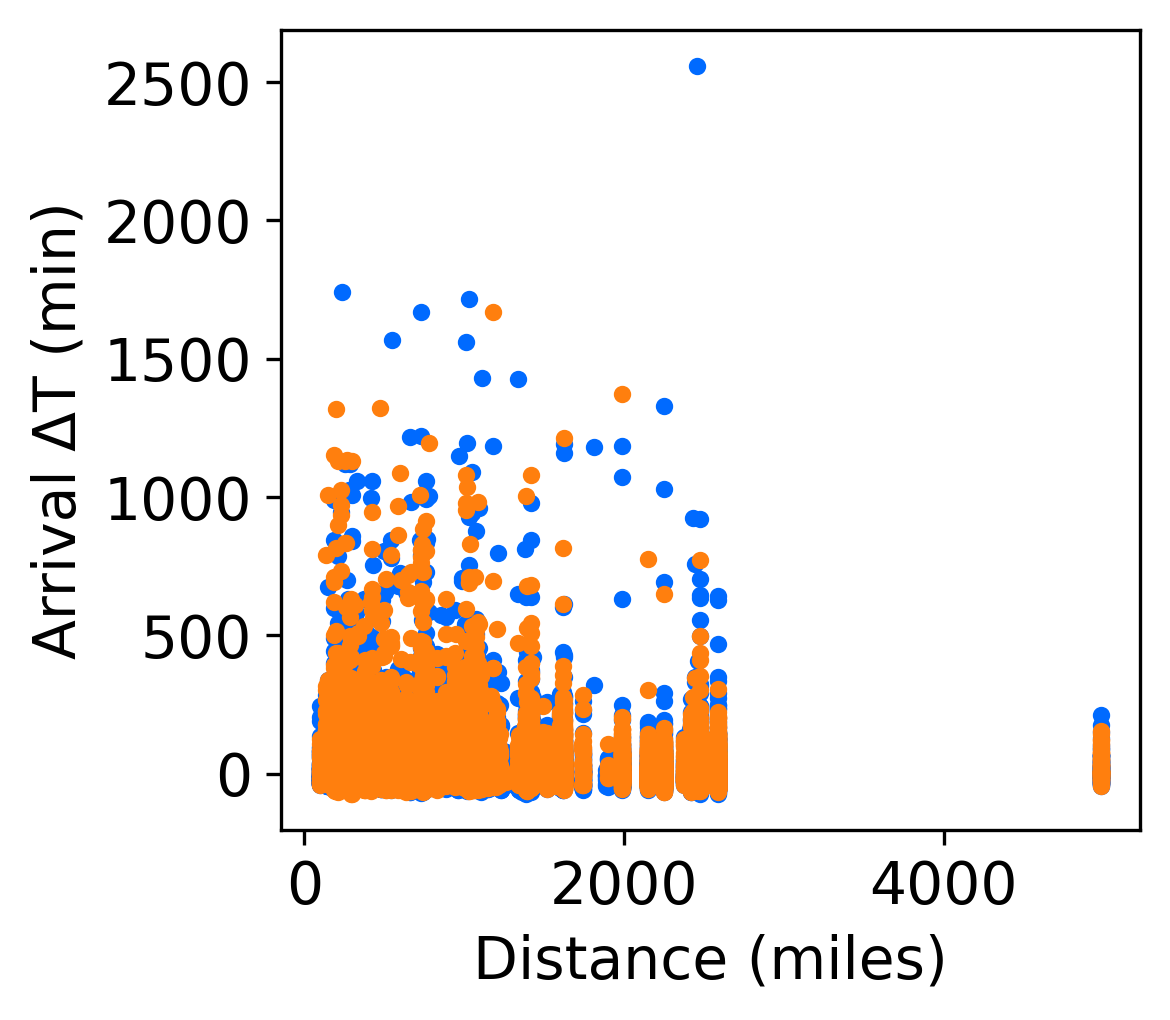} 
        \vspace{-0.7cm}
        \subcaption{\centering Real vs.\ Aug. Synth.}
        \label{fig:operational_4}
    \end{minipage}
    \hfill
    \begin{minipage}{0.23\textwidth} 
        \centering
        \includegraphics[width=\textwidth, trim=0mm 0mm 0mm 0mm, clip]{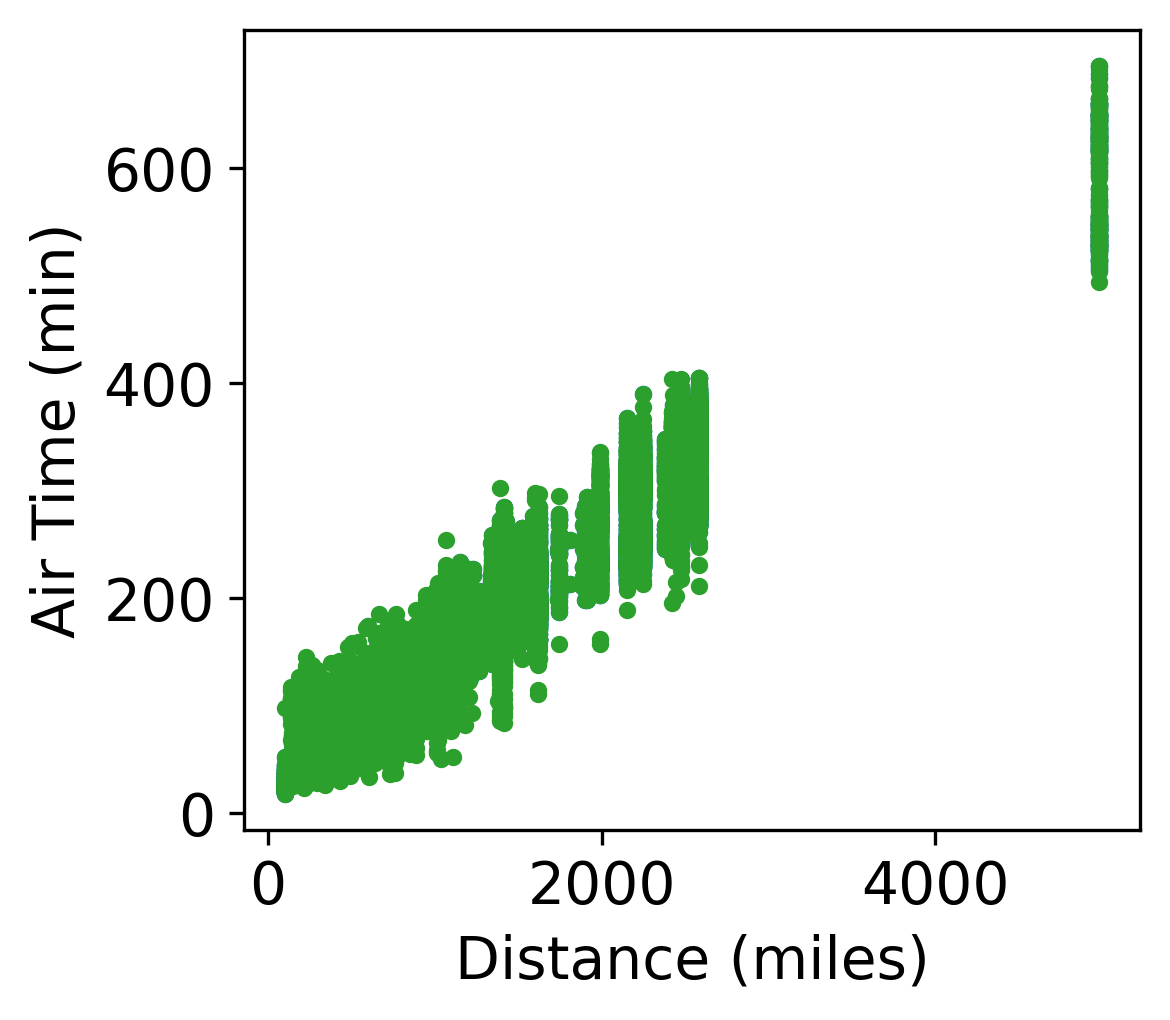} 
        \vspace{-0.7cm}
        \subcaption{\centering Real vs.\ Aug. Real}
        \label{fig:operational_5}
    \end{minipage}
    \hfill 
    \begin{minipage}{0.23\textwidth} 
        \centering
        \includegraphics[width=\textwidth, trim=0mm 0mm 0mm 0mm, clip]{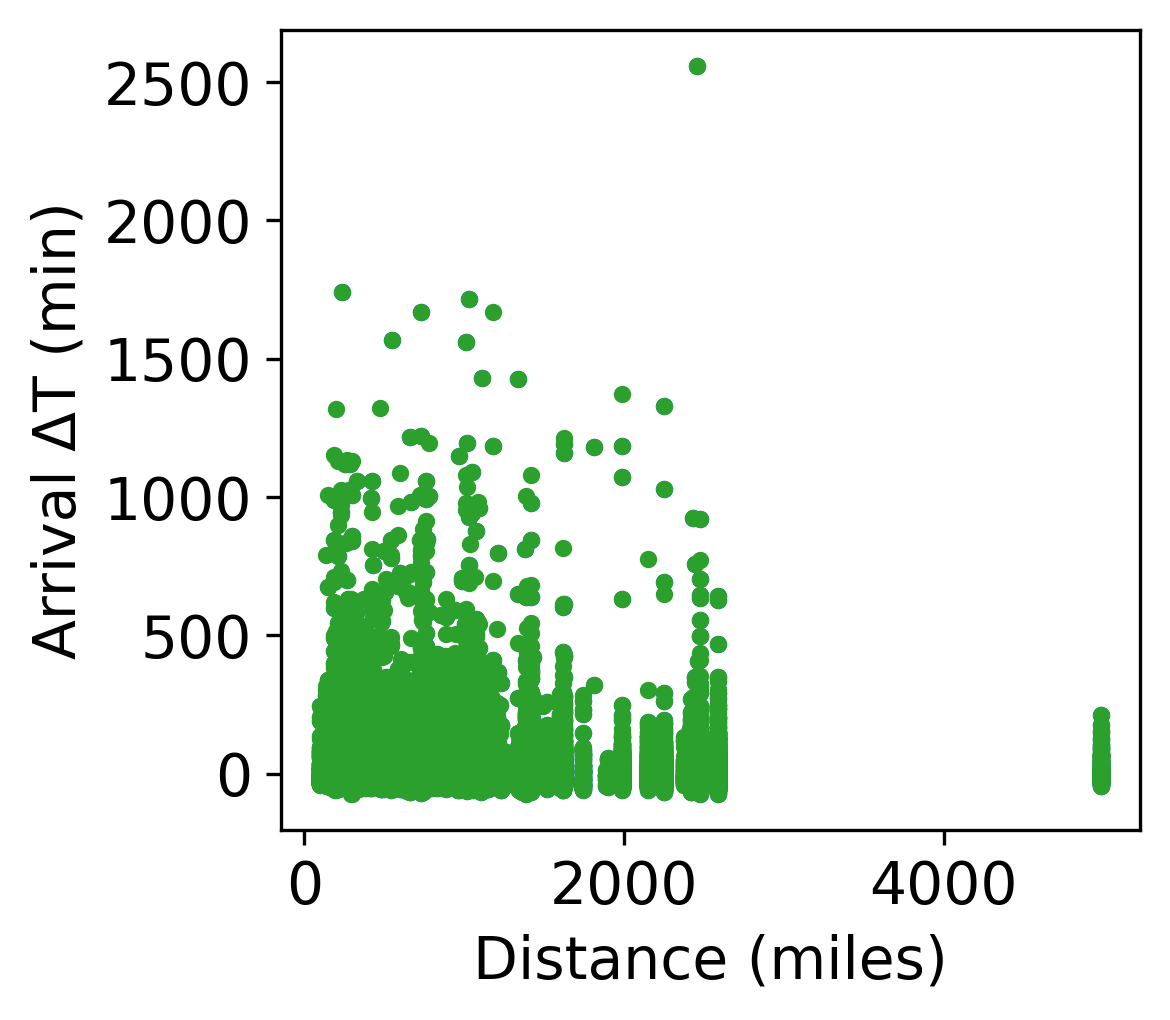} 
        \vspace{-0.7cm}
        \subcaption{\centering Real vs.\ Aug. Real}
        \label{fig:operational_6}
    \end{minipage}
    \caption{\centering Operational correlation in real (blue), na\"{i}ve synthetic (red), augmented synthetic (orange), and augmented real (green) flight records.}
    \label{fig:operational}
\end{figure}

Figure~\ref{fig:operational} presents the pairwise operational correlations for the three comparisons. In all panels, real data are plotted in blue beneath the comparison dataset, allowing direct visual assessment of how closely the synthetic or augmented records adhere to the operational envelope of the real data.

Figures~\ref{fig:operational_1} and~\ref{fig:operational_2} illustrate the limitations of conventional generation. In Figure~\ref{fig:operational_1}, the na\"{i}ve synthetic records fail to cover the extreme air times observed in the real data and include implausible records whose air time--distance pairs fall outside the ranges historically observed for the same routes, violating the operational correlation between flight distance and air time. Figure~\ref{fig:operational_2} reveals an even more pronounced failure: the na\"{i}ve synthetic fails to reproduce the extreme arrival delays present in the real data, with its records concentrated well below the upper tail of the real distribution, which extends to delays of several hundred minutes across all distances.

Figures~\ref{fig:operational_3} and~\ref{fig:operational_4} demonstrate the combined effect of the two anomaly detection layers. Figure~\ref{fig:operational_3} shows the improvement for ``Air Time (min)'': the extreme air times at long-haul distances, absent from Figure~\ref{fig:operational_1}, are now covered through augmentation from $\mathcal{G}_1$. The implausible air time--distance pairs visible in Figure~\ref{fig:operational_1}, whose values fall outside the historical ranges, are absent from Figure~\ref{fig:operational_3}, reflecting the effect of the autoencoder-based cleaning layer, which was trained on the operationally correlated features (``ICAO Origin Airport'', ``ICAO Destination Airport'', ``Air Time (min)'', and ``Distance (miles)'') to learn the empirical operational envelope of historical flight records and remove synthetic records that violate it. In Figure~\ref{fig:operational_4}, the extreme arrival delays are now better represented across the full range observed in the real data, reflecting the direct contribution of $\mathcal{G}_2$, the generative model trained exclusively on the real extreme subset of ``Arrival $\Delta$T (min)''.

Figures~\ref{fig:operational_5} and~\ref{fig:operational_6} show near-perfect adherence to the real data in both correlations, as the augmented real dataset directly incorporates the historical records by construction, augmented with synthetic extremes from $\mathcal{G}_1$ and $\mathcal{G}_2$.

Taken together, these results demonstrate the compounding impact of the two anomaly detection layers of \emph{TailBooster} over conventional generation: the first layer captures and augments the under-represented extremes of user-defined target features, while the second enforces operational validity in a data-driven fashion by learning the empirical operational envelope directly from the historical data. This addresses a limitation consistently observed in our prior work, where synthetic datasets produced by conventional generative models contained operationally implausible records \citep{Aly2025a, Aly2026}.

\subsection{Utility Assessment}
\label{sec:utility_results}
Table~\ref{tab:mae_regression_combined} reports the Mean Absolute Error (MAE) of all six regression models across the four training scenarios for both target features, where all models were trained and tested exclusively on the extreme subsets of each target feature, excluding nominal records as described in Section~\ref{sec:utility_methodology}. Lower MAE indicates better predictive performance on these extreme subsets. Two comparisons are of primary interest: the improvement achieved by training on the Augmented Synthetic dataset ($\mathcal{S}_{\mathrm{aug}}$) over the Na\"{i}ve Synthetic dataset ($\mathcal{S}_{\mathrm{naive}}$), which quantifies the benefit of the dual-layer pipeline over conventional synthetic data generation, and the improvement achieved by training on the Augmented Real dataset ($\mathcal{D}_{\mathrm{aug}}$) over the Real dataset ($\mathcal{D}$), which quantifies the benefit of enriching real historical data with operationally valid synthetic extremes.

For the prediction of extreme ``Air Time (min)'', training on the Augmented Synthetic reduced MAE substantially relative to the Na\"{i}ve Synthetic across all six models, from a range of 19.40--23.58~min to 10.21--12.23~min, representing per-model reductions of approximately 47--49\%. This improvement directly reflects the contribution of the dual-layer pipeline: the extreme-value augmentation from $\mathcal{G}_1$ provides training signal on extreme air times that conventional generation lacks, while the operational cleaning layer ensures that the augmented records respect the empirical air time--distance envelope. Similarly, training on the Augmented Real consistently outperformed training on the Real dataset alone across all models, reducing MAE from 6.34--9.88~min to 2.57--8.59~min, confirming that augmenting the historical record with operationally valid synthetic extremes yields measurable gains in extreme-value prediction even when real data are available.

\begin{table*}[ht]
  \centering
  \footnotesize
  \resizebox{\textwidth}{!}{%
  \begin{tabular}{@{}lcccccccc@{}}
    \toprule
    \multirow{3}{*}{\textbf{Regression Model}}
      & \multicolumn{4}{c}{\textbf{Prediction of ``Air Time (min)''}}
      & \multicolumn{4}{c}{\textbf{Prediction of ``Arrival $\Delta$T (min)''}} \\
    \cmidrule(lr){2-5} \cmidrule(lr){6-9}
      & \multicolumn{4}{c}{\textit{Trained on:}}
      & \multicolumn{4}{c}{\textit{Trained on:}} \\
    \cmidrule(lr){2-5} \cmidrule(lr){6-9}
      & \textbf{Real} & \textbf{Na\"{i}ve Synth.} & \textbf{Aug.\ Synth.} & \textbf{Aug.\ Real}
      & \textbf{Real} & \textbf{Na\"{i}ve Synth.} & \textbf{Aug.\ Synth.} & \textbf{Aug.\ Real} \\
    \midrule
    Random Forest  &  6.34 & 19.40 & 10.34 &  2.57 & 12.09 & 38.89 & 17.35 &  4.49 \\
    XGBoost        &  6.75 & 20.33 & 10.70 &  4.29 & 12.20 & 36.11 & 17.84 &  6.59 \\
    CatBoost       &  6.48 & 20.14 & 10.30 &  6.01 & 11.54 & 34.41 & 14.89 &  8.29 \\
    LightGBM       &  6.45 & 19.88 & 10.21 &  6.23 & 12.13 & 37.22 & 15.96 & 10.73 \\
    SVR            &  9.81 & 23.58 & 12.23 &  8.59 & 17.72 & 41.23 & 20.56 & 15.05 \\
    $k$-NN         &  9.88 & 21.84 & 11.33 &  7.89 & 32.93 & 47.58 & 34.00 & 24.77 \\
    \bottomrule
  \end{tabular}}
  \caption{MAE of regression models trained and tested exclusively on extreme records, for the two target features (lower is better).}
  \label{tab:mae_regression_combined}
\end{table*}

A qualitatively identical pattern was observed for the prediction of extreme ``Arrival $\Delta$T (min)'', though with larger absolute MAE values reflecting the greater variability of this target feature. Training on the Augmented Synthetic reduced MAE relative to the Na\"{i}ve Synthetic from a range of 34.41--47.58~min to 14.89--34.00~min, a reduction of approximately 29--57\% depending on the model. Training on the Augmented Real again outperformed training on the Real dataset alone across all models (4.49--24.77~min vs.\ 11.54--32.93~min).

\begin{figure}[H]
    \centering
    \begin{minipage}{0.23\textwidth} 
        \centering
        \includegraphics[width=\textwidth, trim=0mm 0mm 0mm 0mm, clip]{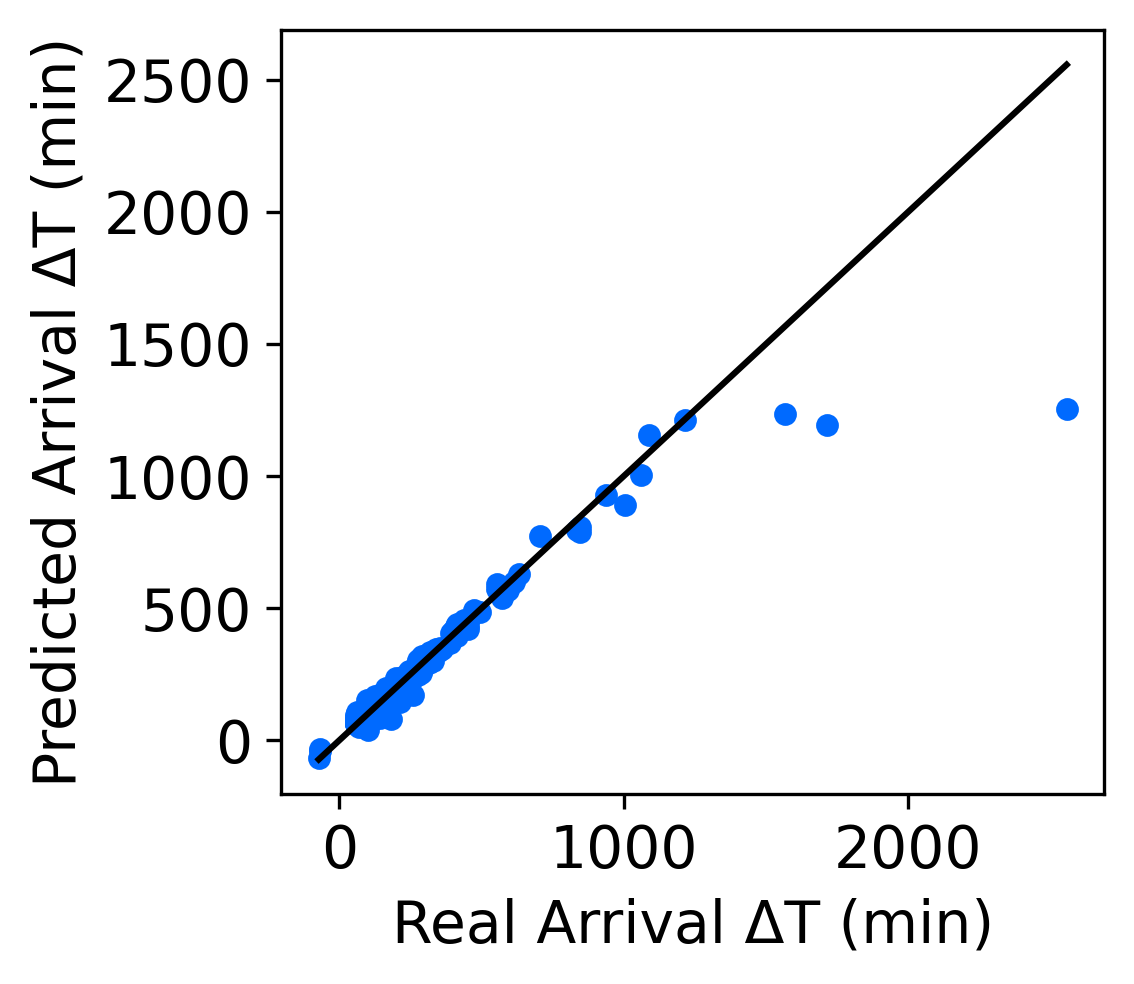} 
        \vspace{-0.7cm}
        \subcaption{\centering Trained on Real}
        \label{fig:utility_1}
    \end{minipage}
    \hfill
    \begin{minipage}{0.23\textwidth} 
        \centering
        \includegraphics[width=\textwidth, trim=0mm 0mm 0mm 0mm, clip]{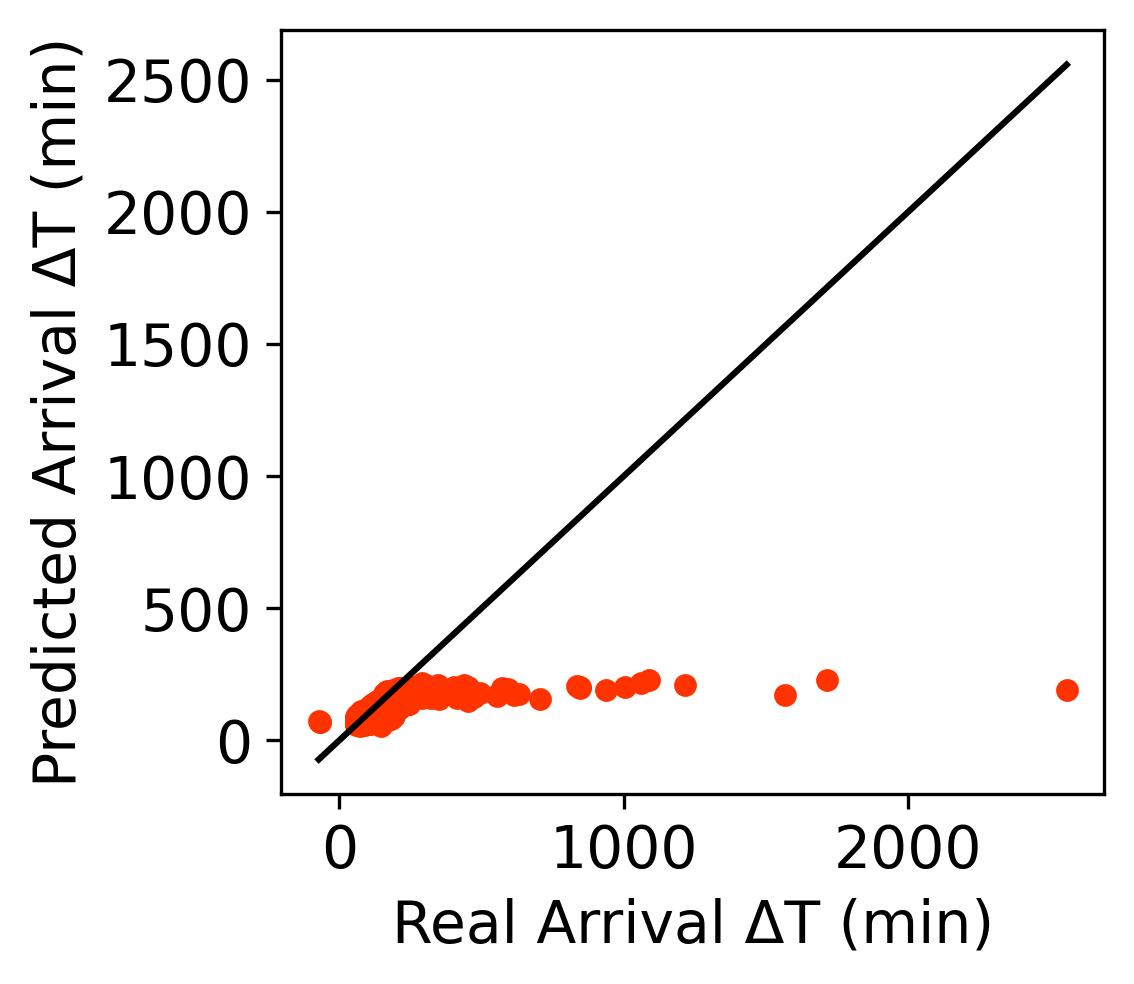} 
        \vspace{-0.7cm}
        \subcaption{\centering Trained on Na\"{i}ve Synth.}
        \label{fig:utility_2}
    \end{minipage}
    \hfill
    \begin{minipage}{0.23\textwidth} 
        \centering
        \includegraphics[width=\textwidth, trim=0mm 0mm 0mm 0mm, clip]{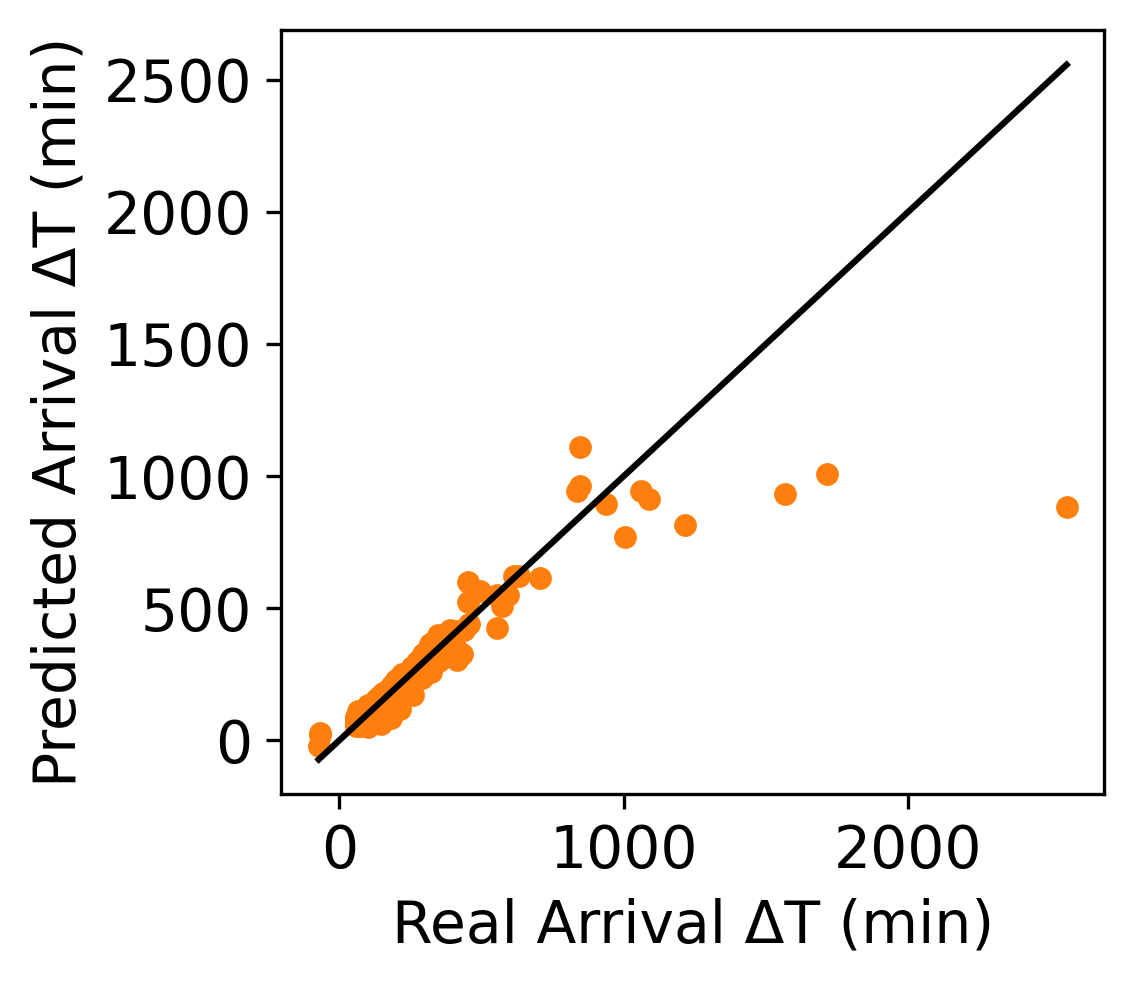} 
        \vspace{-0.7cm}
        \subcaption{\centering Trained on Aug. Synth.}
        \label{fig:utility_3}
    \end{minipage}
    \hfill 
    \begin{minipage}{0.23\textwidth} 
        \centering
        \includegraphics[width=\textwidth, trim=0mm 0mm 0mm 0mm, clip]{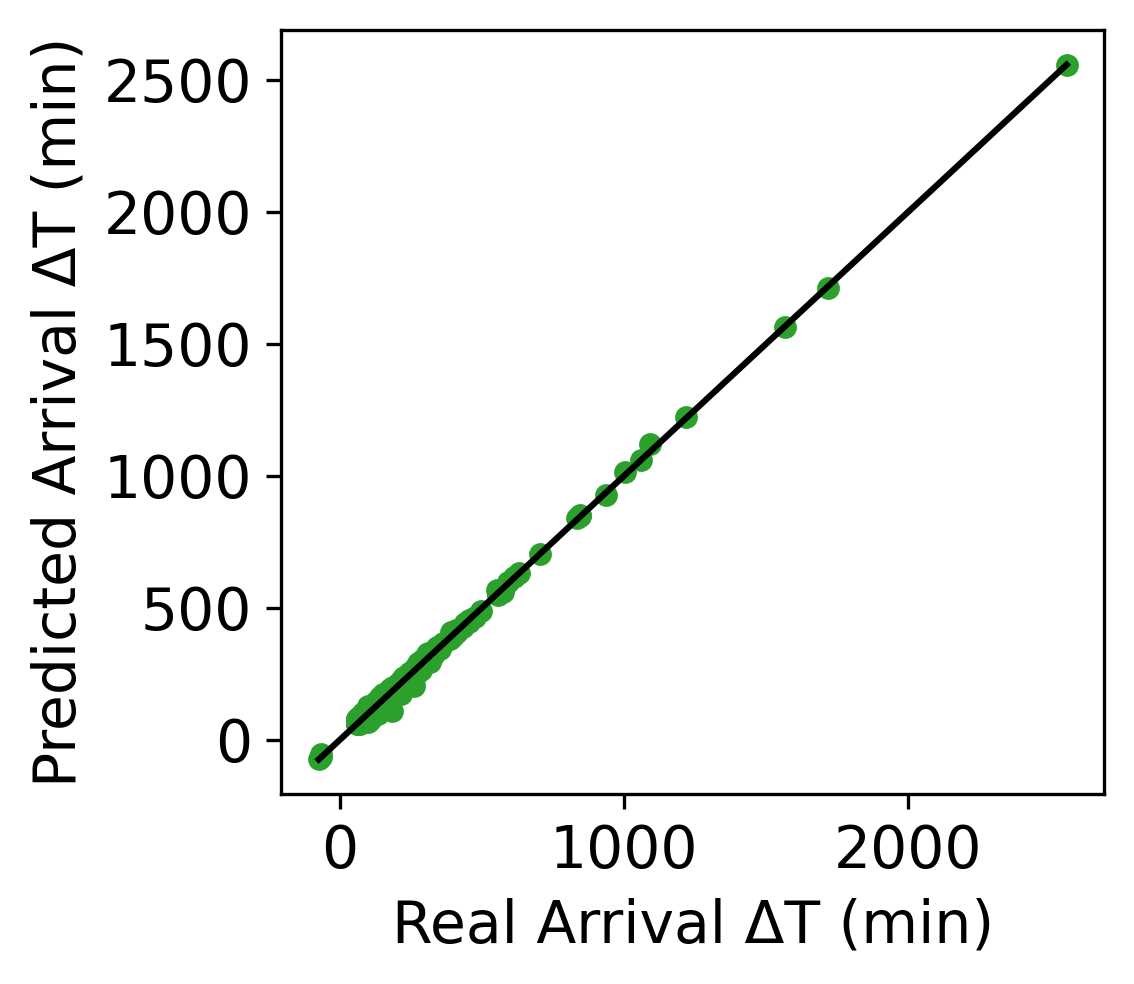} 
        \vspace{-0.7cm}
        \subcaption{\centering Trained on Aug. Real}
        \label{fig:utility_4}
    \end{minipage}
    \caption{\centering Predicted vs.\ real ``Arrival $\Delta$T (min)'' for XGBoost trained on the Real (blue), Na\"{i}ve Synthetic (red), Augmented Synthetic (orange), and Augmented Real (green) datasets. The black diagonal represents perfect prediction.}
    \label{fig:utility}
\end{figure}

Figure~\ref{fig:utility} provides an additional dimension for interpreting the utility results, plotting the real against the predicted ``Arrival $\Delta$T (min)'' for XGBoost trained exclusively on records with extreme ``Arrival $\Delta$T (min)'' from each of the four datasets: Real, Na\"{i}ve Synthetic, Augmented Synthetic, and Augmented Real, and tested on a held-out set of real extremes. Accordingly, nominal records, defined as those with ``Arrival $\Delta$T (min)'' values in the range $[-65.50,\, 58.50]$ minutes, are absent from all four panels. The black diagonal represents perfect prediction; points deviating from this diagonal indicate poor predictive performance.

Figure~\ref{fig:utility_2} illustrates the poor utility of conventionally generated synthetic data for an extreme-value prediction task, with the regression model severely under-predicting extreme arrival delays, its predictions collapsing into a narrow band regardless of the actual extreme value. Figure~\ref{fig:utility_3}, on the other hand, shows the improvement achieved by \emph{TailBooster}: targeted extreme-value augmentation and operational cleaning substantially improve the coverage of extreme predictions relative to the na\"{i}ve case, though some deviation from the diagonal remains at the upper tail.

Figure~\ref{fig:utility_1} shows that a regression model trained exclusively on real historical extremes still struggles to predict the most extreme arrival delays, with points corresponding to the highest real values deviating noticeably from the diagonal. In contrast, Figure~\ref{fig:utility_4} demonstrates that augmenting the real historical extremes with operationally valid synthetic extremes from \emph{TailBooster} and retraining the regression model on the combined dataset substantially closes this gap: the predictions align closely with the diagonal across the full range of extreme values, yielding the best predictive performance of the four scenarios. While Figure~\ref{fig:utility} presents results for predicting extreme ``Arrival $\Delta$T (min)'' only, the same pattern was observed for the prediction of extreme ``Air Time (min)''.

The consistency of these gains across all six models, spanning three distinct regression families, namely tree-based ensembles (Random Forest, XGBoost, CatBoost, and LightGBM, representing both bagging and boosting approaches), kernel-based regression (SVR), and instance-based methods ($k$-NN), confirms that the observed improvements are a property of the augmented data rather than of any particular prediction approach, reinforcing the generalisability of the \emph{TailBooster} framework. These findings were corroborated by RMSE and R\textsuperscript{2} metrics, which, while not reported separately for conciseness, showed trends consistent with the MAE results across all models and both target features.

Together, these results highlight the advantage of \emph{TailBooster} over conventional synthetic data generation methods. The framework is not only valuable for generating synthetic data in settings where access to real data is limited or unavailable, but also in operational environments where historical data exist yet observations in the tails of feature distributions remain under-represented. In such cases, \emph{TailBooster} can enrich the representation of these extreme-value regions, leading to improved predictive performance and more robust modelling of operationally critical events.

\section{Discussion}
\label{sec:discussion} 
As mentioned in Section~\ref{sec:validity_filter}, synthetic flight records must respect the origin--destination airport pairs observed in the historical data; however, deep generative models trained on separate categorical features for ``Origin Airport ID'' and ``Destination Airport ID'' may produce combinations absent from the historical record, representing broken relational structure rather than operationally grounded routes. One established solution is the \texttt{FixedCombinations} constraint provided by the SDV library~\citep{sdv}, which can be used to encode each origin--destination pair as a single route identifier prior to training, thereby reducing two connected features to a single categorical variable that the generative model can learn and reproduce without breaking the origin--destination consistency. The encoding is subsequently reversed after generation, ensuring that the synthetic output contains only route combinations observed in the real data.

\begin{figure}[H]
    \centering
    \begin{minipage}{0.23\textwidth}
        \centering
        \includegraphics[width=\textwidth, trim=0mm 0mm 0mm 0mm, clip]{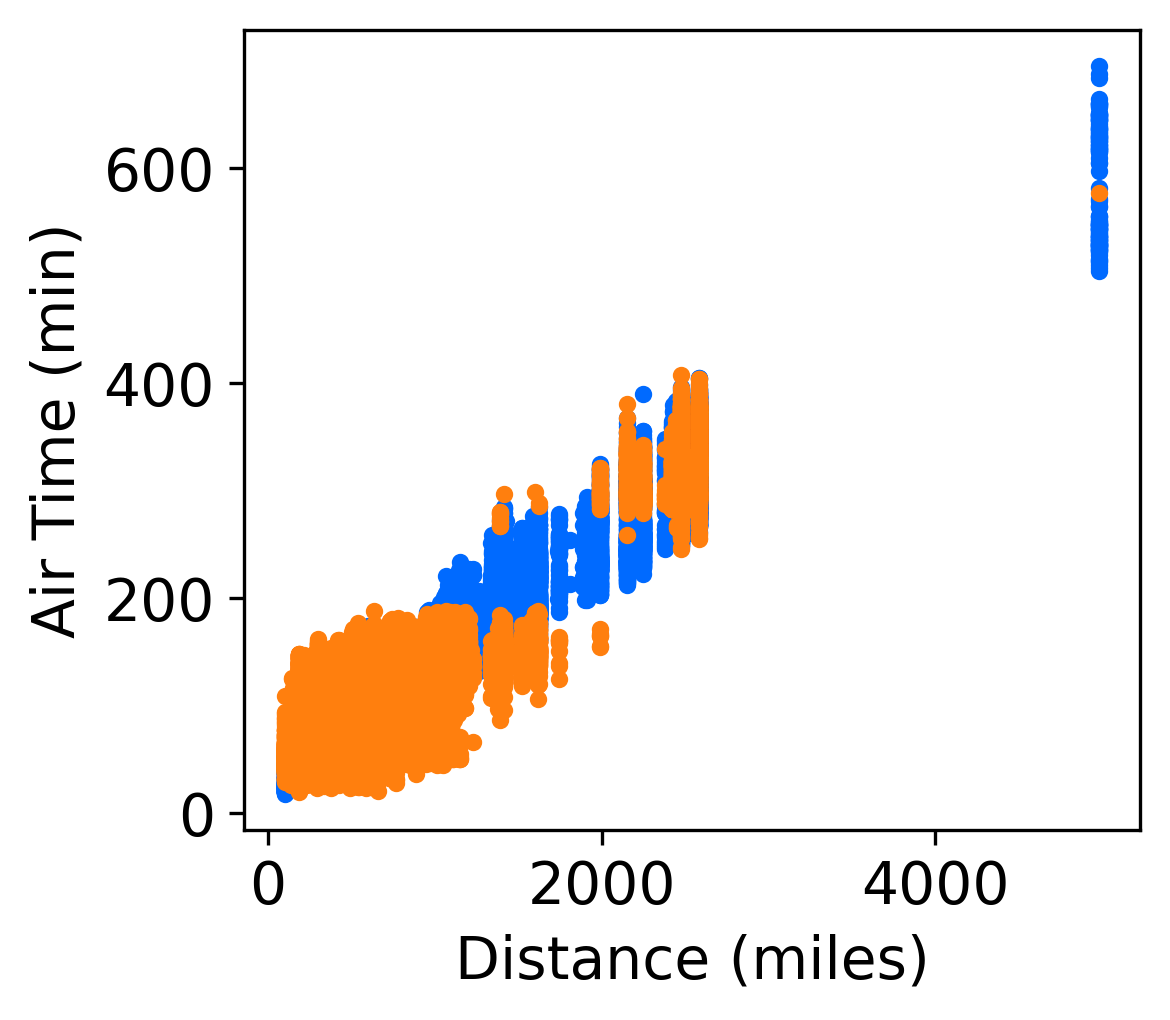}
        \vspace{-0.7cm}
        \subcaption{\centering With the \texttt{FixedCombinations} constraint}
        \label{fig:constraints_1}
    \end{minipage}
    \hfill
    \begin{minipage}{0.23\textwidth}
        \centering
        \includegraphics[width=\textwidth, trim=0mm 0mm 0mm 0mm, clip]{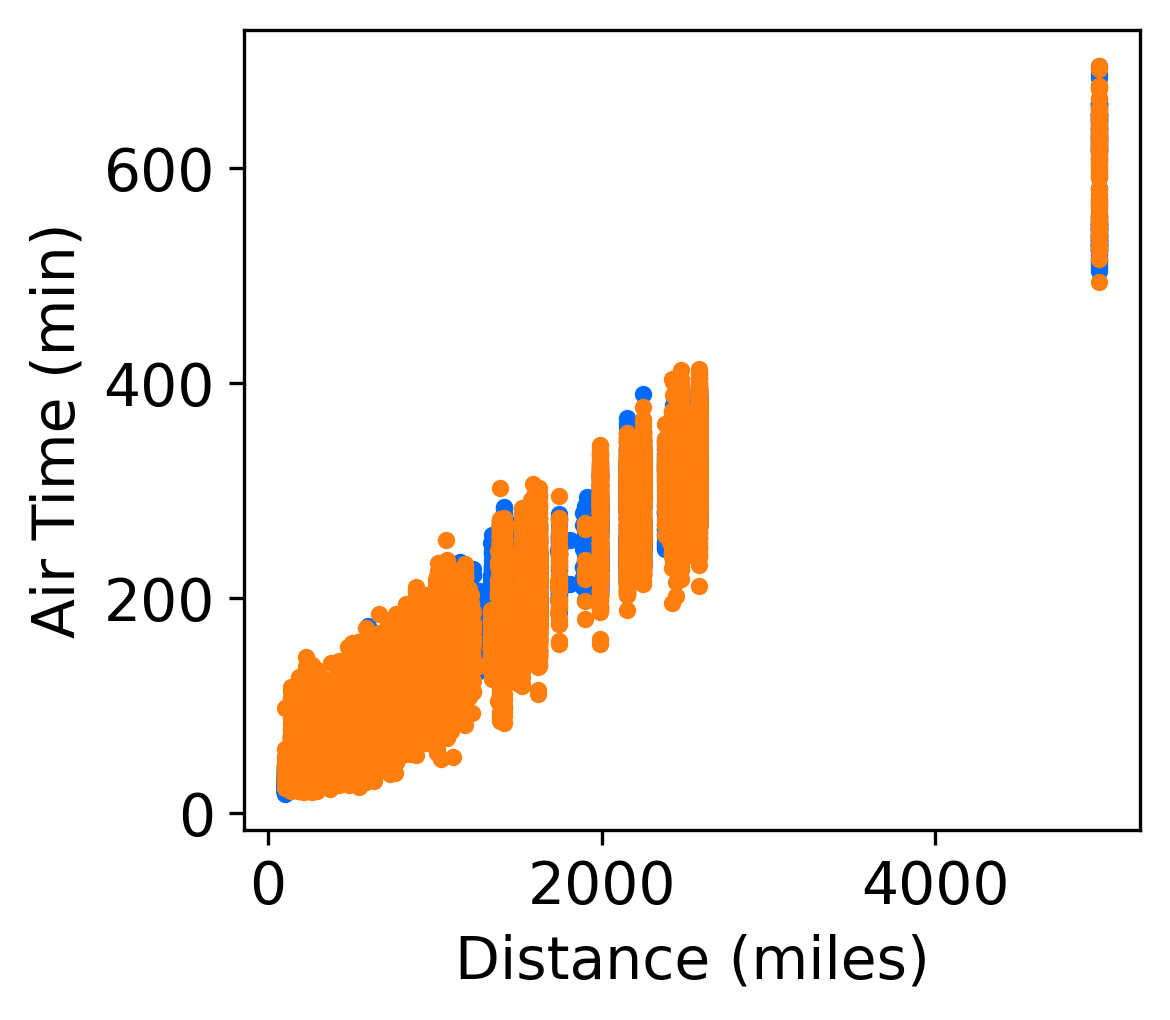}
        \vspace{-0.7cm}
        \subcaption{\centering With the relational validity\\filter}
        \label{fig:constraints_2}
    \end{minipage}
    \caption{\centering Operational correlation for real (blue) vs.\ augmented synthetic (orange) flight records: (a)~distributional collapse resulting from the application of the \texttt{FixedCombinations} constraint; (b)~uniform coverage achieved by replacing the constraint with the post-generation validity filter (Section~\ref{sec:validity_filter}).}
    \label{fig:constraints}
\end{figure}

However, collapsing ``Origin Airport ID'' and ``Destination Airport ID'' into a single route identifier prevents the generative model from associating operational and temporal features with individual airports. Although air times, taxi times, and departure patterns remain present in the training data, the model sees only the combined route token in place of the two airport identifiers and can therefore no longer learn, for instance, that taxi-out time is characteristic of a specific origin airport or that air time scales with the distance profile of a specific destination. As shown in Figure~\ref{fig:constraints_1}, this loss of airport-level associative structure caused distributional collapse, leaving broad regions of the operational space unrepresented.

We therefore preferred to retain ``Origin Airport ID'' and ``Destination Airport ID'' as separate categorical inputs during generative model training and to enforce route validity through the post-generation rejection step described in Section~\ref{sec:validity_filter}. As shown in Figure~\ref{fig:constraints_2}, the resulting augmented synthetic data covers the operational space far more uniformly, without the distributional collapse artefact introduced by the \texttt{FixedCombinations} encoding.

Beyond the design considerations discussed above, \emph{TailBooster} carries practical implications for two distinct practitioner profiles in air transportation. For organisations with access to historical flight records, such as airlines, airports, and air navigation service providers, the framework enriches existing datasets by augmenting the under-represented tails of operationally critical features, directly improving the predictability of extreme events such as severe arrival delays and abnormal air times, as evidenced by the MAE reductions reported in Section~\ref{sec:utility_results} for the Augmented Real dataset over the Real dataset. For researchers and analysts who lack access to real operational data, \emph{TailBooster} provides a mechanism for generating high-quality synthetic data that preserves both nominal patterns and extreme-value structure while ensuring operational validity, as demonstrated by the consistent gains of the Augmented Synthetic dataset over the Na\"{i}ve Synthetic across the key evaluation dimensions in Section~\ref{sec:results}. 

Although demonstrated here in an aviation context, the framework is transferable to any operational domain where the prediction of extreme values of target numerical features is critical. Unlike the domain-constrained approaches reviewed in Section~\ref{sec:related_work}, \emph{TailBooster} enforces operational validity through a fully data-driven mechanism, learning empirical inter-feature correlations directly from historical records rather than encoding them through hand-crafted domain rules, making it applicable to any operational setting where such records are available but governing equations or symbolic constraints are not. 

Finally, the model-agnostic design of the pipeline, in which the TVAE can be substituted with any tabular generative model without modifying any other component, lowers the barrier to adoption and allows practitioners to benefit from future advances in generative modelling without re-engineering the broader framework.

From a computational standpoint, the dominant expense in \emph{TailBooster} lies in training the full-data generative model $\mathcal{G}_0$ on all \num{60767}~flight records across 10~generative input features, and in the hyperparameter optimisation loop, which ran for 100~trials per generative model via the TPE algorithm. Training the extreme-subset models $\mathcal{G}_1$ and $\mathcal{G}_2$ contributes negligible overhead by comparison, as each operates on a substantially smaller dataset confined to the distributional tails of a single target feature (\num{3726}~records for $\mathcal{E}^{(1)}$ and \num{5470}~for $\mathcal{E}^{(2)}$). Similarly, training the autoencoder-based operational cleaning layer is inexpensive: each autoencoder is fitted on a user-defined subset of operationally correlated features drawn from the real historical data, rather than the full feature set (in the present study, 4~features out of 30), and the trained model can subsequently be reused across an unlimited number of generation runs without retraining. All experiments were conducted on a workstation running Ubuntu~22.04.3~LTS, equipped with an AMD Ryzen~9 5950X CPU (16~cores, 32~threads, up to 5.08\,GHz), 128\,GB of RAM, and an NVIDIA GeForce RTX~3060 GPU (12\,GB VRAM).

The present study is nonetheless subject to three limitations. First, the evaluation is conducted on a single dataset comprising U.S.\ domestic New York State flights from January~2023, limiting the temporal and geographic coverage; whether the observed improvements generalise across different seasons, regions, and airport network structures remains to be established. Second, the operational validity dimension is assessed visually through pairwise correlation plots rather than through a quantitative score, which prevents its incorporation into the hyperparameter optimisation objective and limits comparability across studies; formalising it as a quantitative metric, for instance by defining plausibility bounds derived from aircraft performance envelopes or route-level operational statistics, would address both constraints. Third, the extreme subsets used to train the tail-specific generative models are modest in size (Table~\ref{tab:data_size}); a deep generative model trained on few examples has limited diversity in its training signal, which may reduce the variety of synthetic extremes it can produce. This constraint raises a broader question about representativeness: whether the available extreme records are sufficient to capture the full spectrum of real-world tail patterns, and at what point the training subset becomes too sparse for reliable extreme synthesis. This concern would be more pronounced in operational datasets where extreme events are rarer than in the present study.

\section{Conclusions \& Future Work}
\label{sec:conclusions_and_future_work}

This study proposed \emph{TailBooster}, a dual-layer generative framework, combining statistical and deep learning anomaly detection, designed to address two complementary failure modes of conventional deep generative models applied to mixed-type tabular aviation records: the systematic under-representation of distributional tails and the production of operationally invalid synthetic instances. The framework combined IQR-based extreme subset extraction with dedicated TVAE generative models and autoencoder-based operational cleaning, and was evaluated across five distinct dimensions on publicly available U.S.\ domestic flight records.

The three preservation checks confirmed that the augmentation and cleaning processes did not degrade the qualities already achieved by conventional generation: diversity was maintained across all real clusters, and both statistical similarity and fidelity improved relative to the Na\"{i}ve Synthetic baseline. The two primary improvement targets were both met. The data-driven operational cleaning layer markedly reduced the proportion of operationally implausible synthetic records relative to conventional generation. Targeted extreme-value augmentation consistently improved the predictability of extreme events: across six regression models, training on the Augmented Synthetic reduced MAE by approximately 47--49\% on extreme ``Air Time (min)'' and 29--57\% on extreme ``Arrival $\Delta$T (min)'' relative to training on Na\"{i}ve Synthetic data, while training on the Augmented Real consistently outperformed training on real historical records alone, confirming that the improvements are a property of the augmented data rather than of any particular predictive algorithm. 

The results demonstrate that \emph{TailBooster} is beneficial both for practitioners with access to historical flight records, by enriching the representation of extreme-value regions through augmentation of real data with operationally valid synthetic extremes, and for those without such access, by providing augmented synthetic data that substantially outperforms conventionally generated alternatives. Being fully data-driven and generative-model-agnostic, the framework extends naturally to any domain where extreme-event prediction is operationally critical and domain-specific rules governing operational validity are unavailable.

Several directions for future research emerge from these findings. Expanding the evaluation to cover multiple months and a wider range of airports and airline operators would address the limitations in temporal and geographic coverage and enable assessment of whether the framework remains effective under more diverse operational conditions. The framework supports continuous target features extracted via a fixed IQR multiplier; future work could evaluate the sensitivity of results to this threshold and explore its adaptation to features with heavier or more irregular tail behaviour than those examined here, and assess performance across a larger number of target features. Similarly, a sensitivity analysis of the operational cleaning layer with respect to the autoencoder threshold $p$ would offer practitioners clearer guidance on framework configuration across varied operational contexts. In settings where extreme events are sufficiently rare that the extreme subsets $\mathcal{E}^{(k)}$ are too small to support effective deep generative model training, it would be worth investigating whether the tail-specific generators $\mathcal{G}_1, \ldots, \mathcal{G}_{N_{\mathrm{tf}}}$ could be replaced with lightweight resampling methods such as SMOTE or ADASYN. Such a substitution would clarify whether the quality gains of the dual-layer pipeline can be retained while accommodating severe data scarcity in the tails. Finally, in operational settings where governing physical equations are at least partially known, the autoencoder-based cleaning layer could potentially be complemented or replaced by a physics-informed mechanism, such as a physics-informed neural network, that penalises deviation from known governing equations directly. Such an extension would come at the cost of the framework's current transferability---since embedding domain-specific equations would require re-engineering for each new operational setting---trading the generality of the fully data-driven approach for the added precision that known governing laws could provide.

\section*{Code and data Availability}
The code and data used in this study will be made publicly available upon publication at: 
\href{https://github.com/karimyehia92/TailBooster}{https://github.com/karimyehia92/TailBooster}

\bibliographystyle{elsarticle-harv} 
\bibliography{references}

\end{document}